\documentclass[runningheads]{llncs}
\usepackage{graphicx}
\usepackage{comment}
\usepackage{amsmath,amssymb} 
\usepackage{color}
\usepackage{epsfig}
\usepackage{caption}
\usepackage{subcaption}
\usepackage{url}
\usepackage{enumitem}

\begin{document}
\pagestyle{headings}
\mainmatter
\def\ECCVSubNumber{3478}  

\title{CLOTH3D: Clothed 3D Humans} 

\titlerunning{CLOTH3D}
%
\author{Hugo Bertiche\inst{1,2}
\and
Meysam Madadi\inst{1,2}
\and
Sergio Escalera\inst{1,2}
}
\authorrunning{H. Bertiche et al.}
%
\institute{Universitat de Barcelona, Spain 
\and 
Computer Vision Center, Spain
\\
\email{hugo\_bertiche@hotmail.com}
}
\maketitle
\begin{figure}
    \centering
    \includegraphics[width=0.975\textwidth]{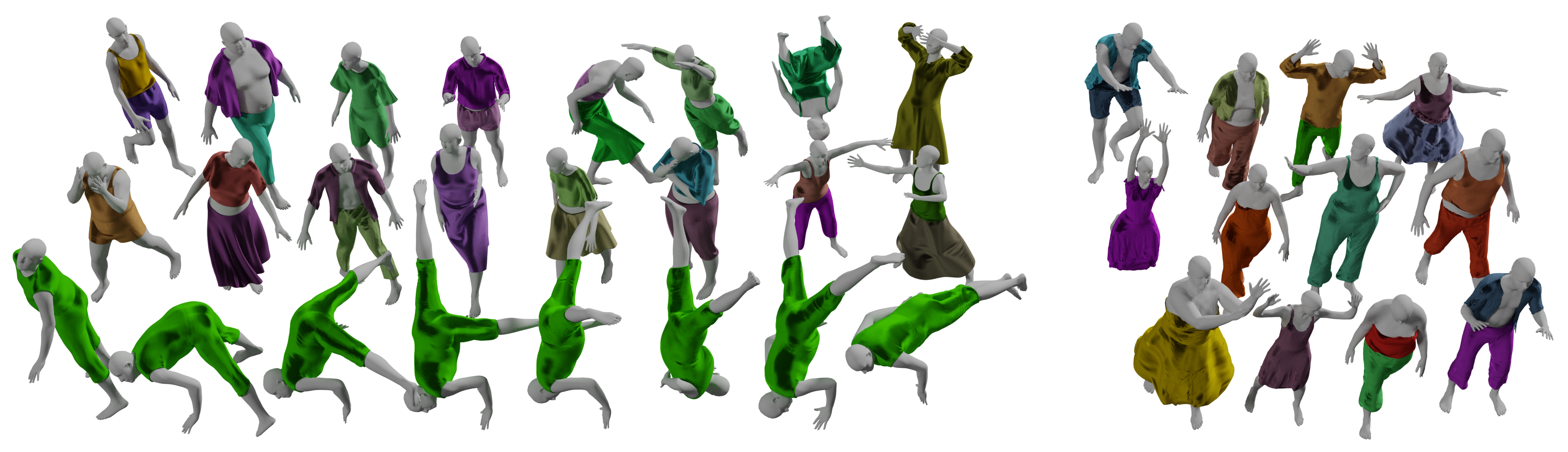}
    \caption[caption]{Left: CLOTH3D\footnotemark{} is the first big scale dataset of animated clothed humans. It contains thousands of different outfits and subjects, high variability of poses and rich cloth dynamics. Right: generated 3D garments with proposed GCVAE.}
    \label{fig:title}
\end{figure}

\begin{abstract}
   We present CLOTH3D, the first big scale synthetic dataset of 3D clothed human sequences. CLOTH3D contains a large variability  on  garment  type,  topology, shape, size, tightness and fabric.  Clothes are simulated on top of thousands of different pose sequences and body shapes, generating realistic cloth dynamics. We provide the dataset with a generative model for cloth generation.    We propose a Conditional Variational Auto-Encoder (CVAE) based on graph convolutions (GCVAE) to learn garment latent spaces.  This allows for realistic generation of 3D garments on top of SMPL model for any pose and shape.
\end{abstract}

\footnotetext{http://chalearnlap.cvc.uab.es/dataset/38/description/}
\section{Introduction}
\label{sec:intro}
    \begin{table}[!t]\setlength\tabcolsep{1pt}
\centering
\renewcommand{\arraystretch}{1}
\makebox[\textwidth][c]{
\begin{tabular}{|c|c|c|c|c|c|c|}
\hline
Dataset & 3DPW\cite{vonMarcard2018} & BUFF\cite{zhang2017detailed} & Untitled\cite{wang2018learning} & 3DPeople\cite{pumarola20193dpeople} & TailorNet\cite{patel2020tailornet} & CLOTH3D \\ \hline
Resolution & 2.5cm & 0.4cm & 1cm & -$^1$ & 1cm & 1cm \\ \hline
Missing & x & \checkmark & x & x & x & x\\ \hline
Dynamics & x & \checkmark & x & x & x & \checkmark \\ \hline
Garments & 18$^2$ & 10$\sim$ 20 & 3$^3$ & High$^4$ & $20$ & 11.3K \\ \hline
Fabrics & x & x & x & x & x & \checkmark \\ \hline
Poses$^5$ & Low & Low & Very low & Low & $1782$ & High \\ \hline
Subjects & 18$^2$ & 6 & 2K & 80 & 9 & 8.5K \\ \hline
Layered & x & x & \checkmark & -$^1$ & \checkmark & \checkmark \\ \hline
\#samples & 51k & 11K & 24K & 2.5M & 55.8k & 2.1M \\ \hline
Type & Real & Real & Synth. & Synth. & Synth. & Synth. \\ \hline
RGB & \checkmark & x & \checkmark & \checkmark & x & x \\ \hline
GT error & 26mm & 1.5-3mm & None & None & None & None \\ \hline
\end{tabular}
}
\caption{CLOTH3D vs. available 3D cloth datasets. $^1$: 3D data includes depth, normal and scene flow maps, but not 3D models. $^2$: 3DPW contains 18 clothed models that can be shaped as SMPL. $^3$: garments of \cite{wang2018learning} are shaped to different sizes. $^4$: Garment variability not specified, nonetheless, authors propose a generation pipeline that can modify template garments into many different sizes. $^5$: poses are strongly related to number of frames, and in \cite{wang2018learning} most samples share the same static pose.}
\label{tab:datasets}
\end{table}

The modelling, recovery and generation of 3D clothes will allow for enhanced virtual try-ons experience, reducing designers and animators workload, or understanding of physics simulations through deep learning, just to mention a few. However, current literature in the modelling, recovery and generation of clothes is almost focused on 2D data~\cite{dong2017multi,lin2015rapid,Pumarola_2019_ICCV,Shin_2019_ICCV}. This is because of two factors. First, deep learning approaches are data-hungry, and nowadays not enough 3D cloth data is available (see Tab. \ref{tab:datasets}). Second, garments present a huge variability in terms of shape, sizes, topologies, fabrics, or textures, among others, increasing the complexity of representative 3D garment generation. 

One could define three main strategies in order to produce data of 3D dressed humans: 3D scans, 3D-from-RGB, and synthetic generation. In the case of 3D scans, they are costly, and at most they can produce a single mesh (human + garments). Alternatively, datasets that infer 3D geometry of clothes from RGB images are inaccurate and cannot properly model cloth dynamics. Finally, synthetic data is easy to generate and is ground truth error free. Synthetic data has proved to be helpful to train deep learning models to be used in real applications~\cite{Nikolenko2019SyntheticDF,Ros_2016_CVPR,varol2017learning}.

In this work, we present CLOTH3D, the first synthetic dataset composed of thousands of sequences of humans dressed with high resolution 3D clothes, see Fig.\ref{fig:title}. CLOTH3D is unique in terms of garment, shape, and pose variability, including more than 2 million 3D samples. We developed a generation pipeline that creates a unique outfit for each sequence in terms of garment type, topology, shape, size, tightness and fabric. While other datasets contain just a few different garments, ours has thousands of different ones. On Tab. \ref{tab:datasets} we summarize features of existing datasets and CLOTH3D.

Additionally, we provide a baseline model able to generate dressed human models. Similar to \cite{alldieck2018detailed,ma2019dressing,yang2018analyzing} we encode garments as offsets connecting skin to cloth, using SMPL\cite{SMPL:2015} as human body model. This yields an homogeneous dimensionality on the data. As in~\cite{pons2017clothcap}, we use a segmentation mask to extract the garment by removing body vertices. In our case, the mask is predicted by the network.
We propose a Conditional Variational Auto-Encoder (CVAE) based on graph convolutions \cite{bronstein2017geometric,defferrard2016convolutional,ma2019dressing,niepert2016learning,wu2019comprehensive,yuan2019mesh} (GCVAE) to learn garment latent spaces. This later allows for the generation of 3D garments on top of SMPL model for any pose and shape (right on Fig.\ref{fig:title}).

\section{Related Work}
\label{sec:sota}
    \textbf{3D Garment Datasets.} Current literature on 3D garment lacks on large public available datasets. One strategy to capture 3D data is through \textbf{3D scans}. The BUFF dataset \cite{zhang2017detailed} provides high resolution 3D scans, but few number of subjects, poses and garments. Furthermore, scanning techniques cannot provide layered models (one mesh for the body and one for each garment) and often one can find regions occluded at scanning time, meaning missing vertices or corrupted shapes. The work of \cite{pons2017clothcap} proposed a methodology to segment scans to obtain layered models. Authors of \cite{yu2019simulcap} combined 3D scans with cloth simulation fitting at each frame to deal with missing vertices. Similarly, \cite{bhatnagar2019multi} provided a dataset from 3D scans. However, the amount of samples is in the order of a few hundreds. The 3DPW dataset~\cite{vonMarcard2018} is not focused on garments, but rather on pose and shape in-the-wild. The authors proposed a modified SMPL \textbf{parameterized model} for each outfit (18 clothed models), which, as SMPL, can be shaped and posed. Nevertheless, resolution is low and posing is through rigid rotations. Therefore, cloth dynamics are not represented. The dataset of \cite{wang2018learning} is synthetically created through \textbf{physics simulation}, with three different garment types: tshirt, skirt and kimono. They propose an automatic garment resizing based on real patterns, but provide only static samples on few different poses. The work of \cite{patel2020tailornet} also includes a synthetic dataset obtained through simulation of $20$ combinations of different garment styles and body shapes into $1782$ static poses. Finally, 3DPeople dataset \cite{pumarola20193dpeople} is the most comparable to ours in terms of scale, but has significant differences w.r.t. CLOTH3D. On one hand, this dataset has been designed specifically for computer vision. Data are given as \textbf{multi-view images} (RGB, depth, normal and scene flow), there are no 3D models. On the other hand, the garments are rigged models, so there is no proper cloth dynamics. And lastly, source pose data is sparse, $70$ pose sequences with an average length of $110$ frames. Our CLOTH3D dataset aims to overcome previous datasets issues. We automatically generate garments to obtain a huge variability on garment type, topology, shape, size, tightness and fabric. Afterwards, we simulate clothes on top of thousands of different pose sequences and body shapes. Tab.\ref{tab:datasets} shows a comparison of features for existing datasets and ours. In CLOTH3D we focus on sample variability (garments, poses, shapes), containing realistic cloth dynamics. 3DPW and 3DPeople sequences are based on rotations on rigged models, datasets of \cite{patel2020tailornet,wang2018learning} contain static poses only, and BUFF has very few and short sequences. Moreover, none other provides metadata about fabrics, which has a strong influence on cloth behaviour. Similarly, the scarcity of these datasets implies low variability on garments, poses and subjects. Finally, note how only synthetic datasets provide with layered models and have no annotation error.

\textbf{3D Garment Generation.} Current works in 3D clothing focus on the generation of dressed humans. We split related work into non-deep and deep-learning approaches. Regarding \textbf{non-deep learning}, the authors of \cite{guan2012drape} proposed a data-driven model that learns deformations from template garment to garment fitted to the human body, shaped and posed. They factorize deformations into shape-dependant and pose-dependant by training on rest pose data first, and later on posed bodies. Transformations are learnt per triangle, and thus it yields inconsistent meshes that need to be reconstructed. The data-driven model of \cite{pons2017clothcap} is able to recover and retarget garments from 4D scan sequences relying on masks to separate body and cloth. Authors propose an energy optimization process to identify underlying body shape and garment geometry, later, cloth displacements w.r.t. body are computed and applied to new body shapes. This means information such as wrinkles is "copied" to new bodies, which produces valid samples but cannot properly generate its variability. Regarding \textbf{deep learning} strategies, the work of \cite{gundogdu19garnet} deals with body and garments as different point clouds through different streams of a network, which are later fused. They also use skin-cloth correspondences for computing local-features and losses through nearest neighbour. The works of \cite{alldieck2018detailed,ma2019dressing,patel2020tailornet,yang2018analyzing} consider encoding clothes as offsets from SMPL body model with different goals. In \cite{ma2019dressing} authors propose a combination of graph VAE and GAN to model SMPL offsets into clothing.  Similarly, in \cite{patel2020tailornet}, authors propose encoding garments as SMPL offsets and topology as a subset of SMPL vertices, later, they learn two models for low and high frequency details  which effectively generate realistic wrinkles on the garments. In \cite{wang2018learning,yang2018analyzing} a PCA decomposition is used to reduce clothing space. In \cite{alldieck2019tex2shape,lahner2018deepwrinkles}, authors register garments to low resolution meshes (garment templates and SMPL respectively), to later use UV normal maps to represent high-frequency cloth details (wrinkles). Authors of \cite{santesteban2019learning} propose learning Pose Space Deformation models for template garments by training deep models instead of SVD  (as SMPL). The work of \cite{wang2019learning} presents a template garment autoencoder where latent spaces are disentangled into motion and static properties to realistically interpolate into 3D keyframes. Similar to previous approaches, our proposed methodology also encodes clothes as SMPL offsets. Nevertheless, the assumption that garments follow body topology does not hold for skirts and dresses. In this sense, we propose a novel body topology specific for those cases. Additionally, our model predicts garment mask along offsets to generate layered models.

\section{Dataset}
\label{sec:dataset}
    CLOTH3D is the first big scale dataset of 3D clothed humans. The dataset is composed of 3D sequences of animated human bodies wearing different garments. Fig. \ref{fig:title} depicts a sequence (first row) and randomly sampled frames from different sequences. Samples are layered, meaning each garment and body are represented by different 3D meshes. Garments are automatically generated for each sequence with randomized shape, tightness, topology and fabric, and resized to target human shape. This process yields a unique outfit for each sequence. It contains over $7000$ non-overlapping sequences of $300$ frames each at $30$fps, yielding a total of 2.1M samples. As seen in Tab. \ref{tab:datasets}, garment and pose variability is scarce in available datasets, and CLOTH3D aims to fill that gap. To ensure garment type balance, given that females present higher garment variability, we balance gender as 2:1 (female:male). Finally, for validation purposes, we split the data in 80\% sequences as training and 20\% as test. Splitting by sequences ensures no garment, shape or pose is repeated in training and test. 

\begin{figure}[!t]
    \centering
    \includegraphics[scale=.13]{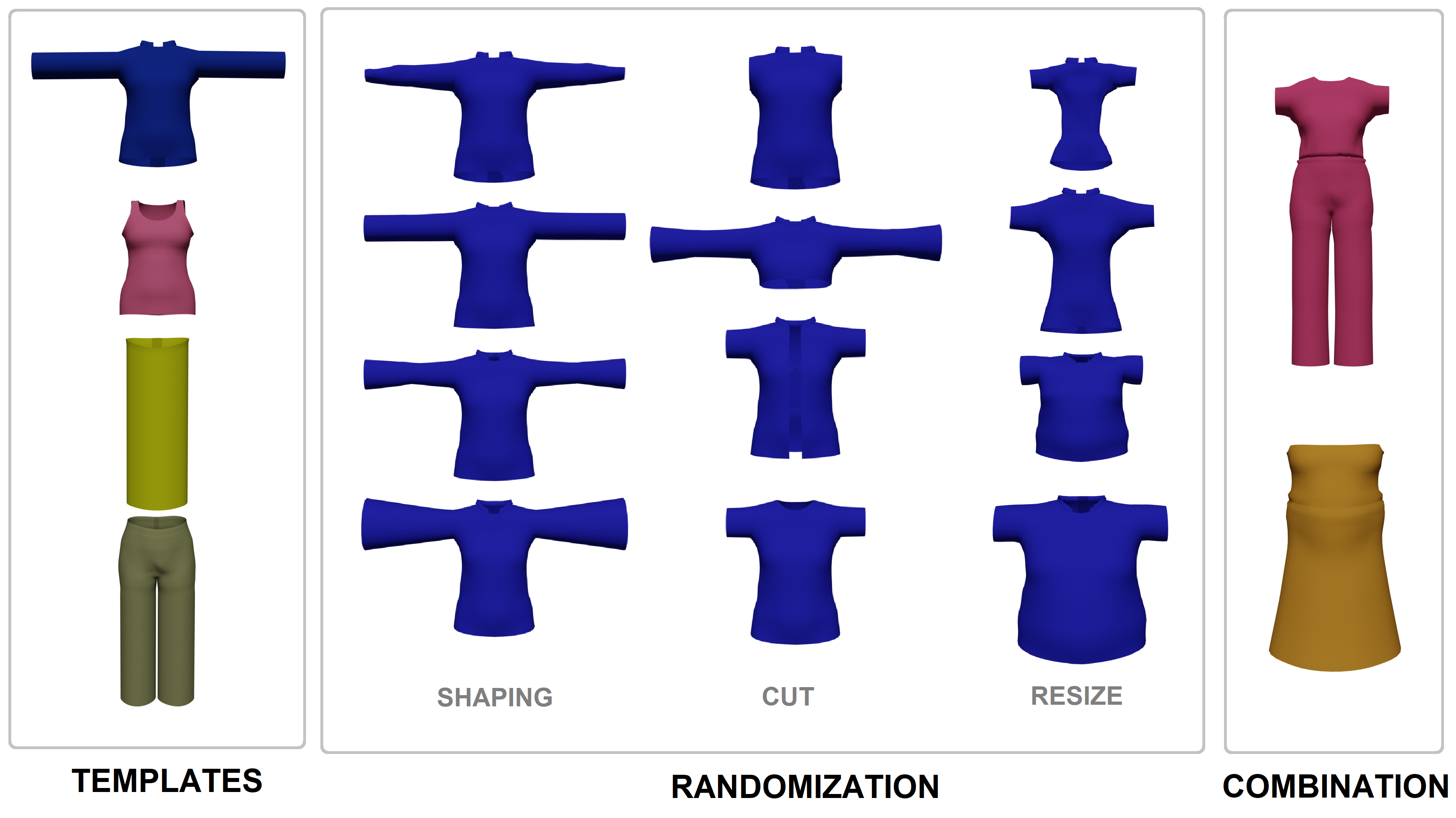}
    \caption{Unique outfit generation pipeline. First, one upper-body and lower-body garment template is selected. Then, garments are individually shaped, cut and resized. Finally, garments might be combined into a single one.}
    \label{fig:cloth3d_pipeline}
\end{figure}

The data generation pipeline starts with sequences of human bodies in 3D. Human pose data source is \cite{CMUMocap}, later transformed to volumetric bodies through SMPL \cite{SMPL:2015}. These sequences might present body self-collisions which will hinder cloth simulation, not only on affected regions, but also in global garment dynamics. We automatically solve collisions or reject these samples. Human generation process is described in subsec. \ref{subsec:humans}. Later, we generate unique outfits for each sequence. We start from a few template meshes which are randomly shaped, cut and resized to generate a unique pair of garments for each sample, with the possibility to be combined into a single full-body garment. Fig. \ref{fig:cloth3d_pipeline} shows the generation process, which is also detailed in subsec. \ref{subsec:garment_generation}. Finally, once human sequence and outfit are done, we use a physics based simulation to obtain the garment 3D sequences. Simulation details are described in subsec. \ref{subsec:simulation}.

\subsection{Human 3D Sequences}
\label{subsec:humans}

\textbf{SMPL.} It is a parametric human body model which takes as input shape $\beta\in \mathbb{R}^{10}$ and pose $\theta\in \mathbb{R}^{24\times3}$ to generate the corresponding mesh with $6890$ vertices. We use this model to generate animated human 3D sequences. We refer to \cite{loper2015smpl} for SMPL details. To generate animated bodies, we need a source of valid sequences of SMPL pose parameters $\theta\in \mathbb{R}^{f\times24\times3}$. We take such data from the work of \cite{varol2017learning}, where pose is inferred from CMU MoCap data \cite{CMUMocap} following the methodology proposed at \cite{loper2014mosh}. These pose data come from around 2600 sequences of 23 different actions (dancing, playing, running, walking, jumping, climbing, etc.) performed by over 100 different subjects. SMPL shape deformations are linearly modeled through PCA. To obtain a balanced dataset we uniformly sample shape within range $[-3, 3]$ for each sequence.

\begin{figure}[t!]
    \centering
    \includegraphics[width=0.65\textwidth]{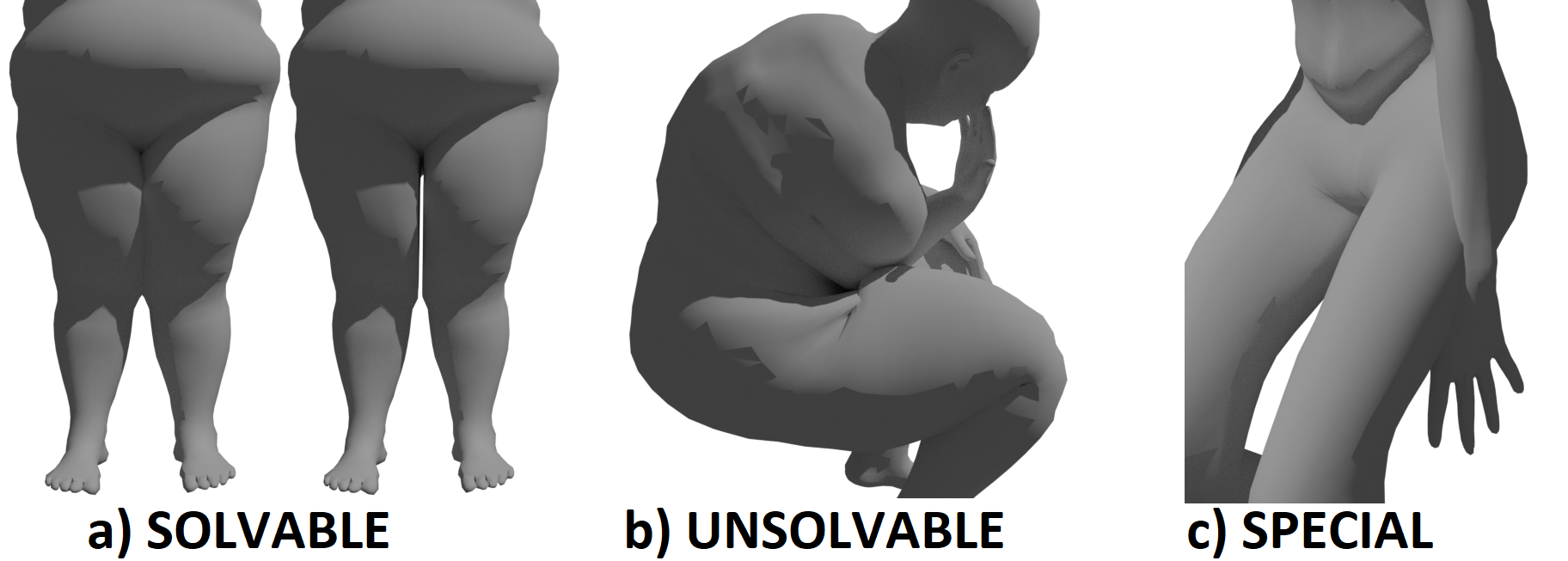}
    \caption{Types of self-collision: a) collided vertices can be linearly separated with the aid of a body part segmentation, b) no trivial solution, we reject this kind of sample, c) correct simulation might be possible if forearm is removed.}
    \label{fig:selfcollision}
\end{figure}

\textbf{Self-collision.} Body collides with itself for certain combinations of pose and shape parameters. Intersection volumes create regions where simulated repel forces are inconsistent, corrupting global cloth dynamics. We classify these collisions in three generic cases. Solvable Fig.\ref{fig:selfcollision}(a): small intersection volumes near joints, specially armpits and crotch. We use SMPL body parts segmentation to linearly separate the collided vertices to permit a correct simulation. Separation space is $4$mm so that a folded cloth can fit. Unsolvable Fig.\ref{fig:selfcollision}(b): big intersection volumes or incompatible intersections (e.g.: arm vs. leg). We reject or re-simulate with thinner body. Special cases Fig.\ref{fig:selfcollision}(c): removing hands, forearms or arms for short-sleeved upper-body and lower-body garments significantly increases the amount of valid samples. This requires manual supervision. Self-collision solution is not stored, hence, if collided vertices change significantly, garments might present interpenetration w.r.t. unsolved body. Only small intersected volumes are corrected and the rest are rejected (or simulated with thinner body). The goal of self-collision solving is to avoid invalid cloth dynamics. Accurate, realistic solving of soft-body self-collision is out of the scope of this work.

\subsection{Garment Generation}
\label{subsec:garment_generation}

\textbf{Garment Templates.} Generation starts with a few template garments for each gender. Garments can be classified in upper-body and lower-body. Lower-body can be further split into trousers and skirts. These three categories, and combinations between them, encompass almost any day-to-day garment. Template garments have been manually created by designers from real patterns and are: t-shirt, top, trousers and skirt.

\textbf{Shaping.} On sleeves, legs and skirt, we find a significant shape variability. It is possible to define them as cylinders of variable width around certain axes: along arms for sleeves, legs for trousers and vertical body axis for skirt. For sleeves and legs, width will be constant or decreasing while moving towards wrist/ankle, and beyond a randomly sampled point along its axis, it might start increasing (widening). For skirts,
width always increases, from waist to bottom. Rate of width decrease/increase is uniformly sampled within ranges empirically set per garment. 
More formally:
\begin{equation}
    W(x) = \alpha_1 x + \alpha_2 \max(0, x - x_{offset}) + W_0,
\end{equation}
where $x$ is position along axis ($0$ at shoulder/hips), $W(x)$ is width at position $x$, $W_0$ is width at $x=0$, $x_{offset}$ is a uniformly sampled point along the axis and $\alpha_1$ and $\alpha_2$ are constants empirically defined for each garment. For t-shirts and trousers, $\alpha_1<0<\alpha_2$. For skirts, $\alpha_1 > \alpha_2 = 0$.

\textbf{Cut.} Template garments cover most of the body (long sleeves, legs and skirt). At this generation step, garments are cut to increase variability on length and topology. Cuts are along arms, legs and torso. Plus, upper-body garments have specific cuts to generate different types of garments (e.g., t-shirt, shirt, polo).

\textbf{Resizing.} Garments are resized to random body shapes. It is safe to assume that size variability on garments is similar to body shape variability. Following this reasoning, SMPL shape displacements are transferred to garments by nearest neighbour. Nevertheless, this process is noisy and human body details are transferred to garment. To address these issues, an iterative Laplacian smoothing is applied to shape displacements, removing noise and filtering high frequency body details, while preserving the geometry of the original garment. On SMPL, first and second shape parameters correspond to global human size and overall fatness. Knowing this, garments are resized to a different target shape. This new shape has two offsets at first and second parameters, the garment tightness $\gamma\in\mathbb{R}^2$. These offsets on garment resizing will generate loose or tight variability. As tighter garments present less dynamics and complexity, we bias the generator towards loose clothes by sampling tightness on the range $[-1.5, 0.5]$.

\textbf{Jumpsuits and Dresses.} Full-body garments can be generated by combining upper-body and lower-body garments. After generating the clothes individually, a final step automatically sews them together.

\subsection{Simulation}
\label{subsec:simulation}

Cloth simulation is performed on Blender, an open source 3D creation suite. Blender's cloth physics, as it is in version 2.8, has been implemented with state-of-the-art algorithms based on mass-spring model. The simulation performs $420-600$ steps per second, depending on the complexity of the garment.

\textbf{Fabrics.} Changing the parameters of the mass-spring model allows simulation of different fabrics. Blender provides different presets for \textit{cotton}, \textit{leather}, \textit{silk} and \textit{denim}, among others. These four fabrics have been used for the creation of the dataset. Upper-body garments might be cotton or silk, while the rest of the garment types can be any of those fabrics. Different fabrics produce different dynamics and wrinkles on simulation time.

\textbf{Elastics.} At simulation time, sleeves and legs have a $50\%$ chance each of presenting an elastic behaviour at their ends, also at waist on full-body garments.

\subsection{Additional dataset statistics}
\label{sec:cloth3d_stats}
Tab.\ref{tab:action_stats} shows the CLOTH3D statistics in terms of action labels by grouping them into generic categories. Note that original data action label is very heterogeneous, specific and incomplete. These labels are gathered from CMU MoCap dataset. We observe a high density on \textit{Walk}, but it is important to note that this gathers many different sub-actions (walk backwards, zombie walk, walk stealthily, ...) as many other action labels do. Additionally, most of these actions were performed by different subjects, which implies an increase in intra-class variability. The label 'others' contains all action labels that cannot be included in any of the categories plus all the missing action labels.

\begin{table}[!t]
\centering
\makebox[\textwidth][c]{
\begin{tabular}{|c|c|c|c|c|c|c|c|c|c|c|c|c|}
\hline
Walk    & Animal  & Fight  & Jump   & Run    & Sing   & Wait   & Swim   & Story  & Sports & Dance  & Yoga   & Spin   \\ \hline
27.49\% & 10.79\% & 4.38\% & 2.78\% & 2.49\% & 2.38\% & 2.31\% & 1.97\% & 1.70\% & 1.63\% & 1.37\% & 1.01\% & 0.90\% \\ \hline
\end{tabular}
}
\makebox[\textwidth][c]{
\begin{tabular}{|c|c|c|c|c|c|c|c|c|c|c|c|}
\hline
Exercise & Climb  & Carry  & Stand  & Wash   & Balancing & Trick  & Sit    & Interact & Drink  & Pose   & Others  \\ \hline
0.84\%   & 0.71\% & 0.67\% & 0.66\% & 0.63\% & 0.54\%    & 0.51\% & 0.28\% & 0.20\%   & 0.14\% & 0.14\% & 33.48\% \\ \hline
\end{tabular}
}
\caption{CLOTH3D statistics per action label.}
\label{tab:action_stats}
\end{table}

\section{Dressed Human Generation}
\label{sec:methodology}
    This section presents the methodology for deep garment generation. As \cite{alldieck2018detailed,ma2019dressing,patel2020tailornet,yang2018analyzing}, data dimensionality and topology is fixed by encoding it as body offsets. In addition, by masking body vertices we represent different garment types and separate them from the body, e.g. in a similar fashion to \cite{patel2020tailornet,pons2017clothcap}. To compute ground truth offsets, a body-to-garment matching is needed. A dedicated algorithm for this task should be able to correctly register skirt-like garments which have a different topology than the body. In sec. \ref{sec:data} we explain details of our data pre-processing. Our proposed model is a Graph Conditional Variational Auto-Encoder (GCVAE). By conditioning on available metadata (pose, shape and tightness), we learn a latent space encoding specific information about garment type and its dynamics (details are given in sec. \ref{sec:gcvae}). Fig. \ref{fig:pipeline} illustrates the proposed model.

\begin{figure}[!t]
    \centering
    \includegraphics[width=.65\textwidth,height=3cm]{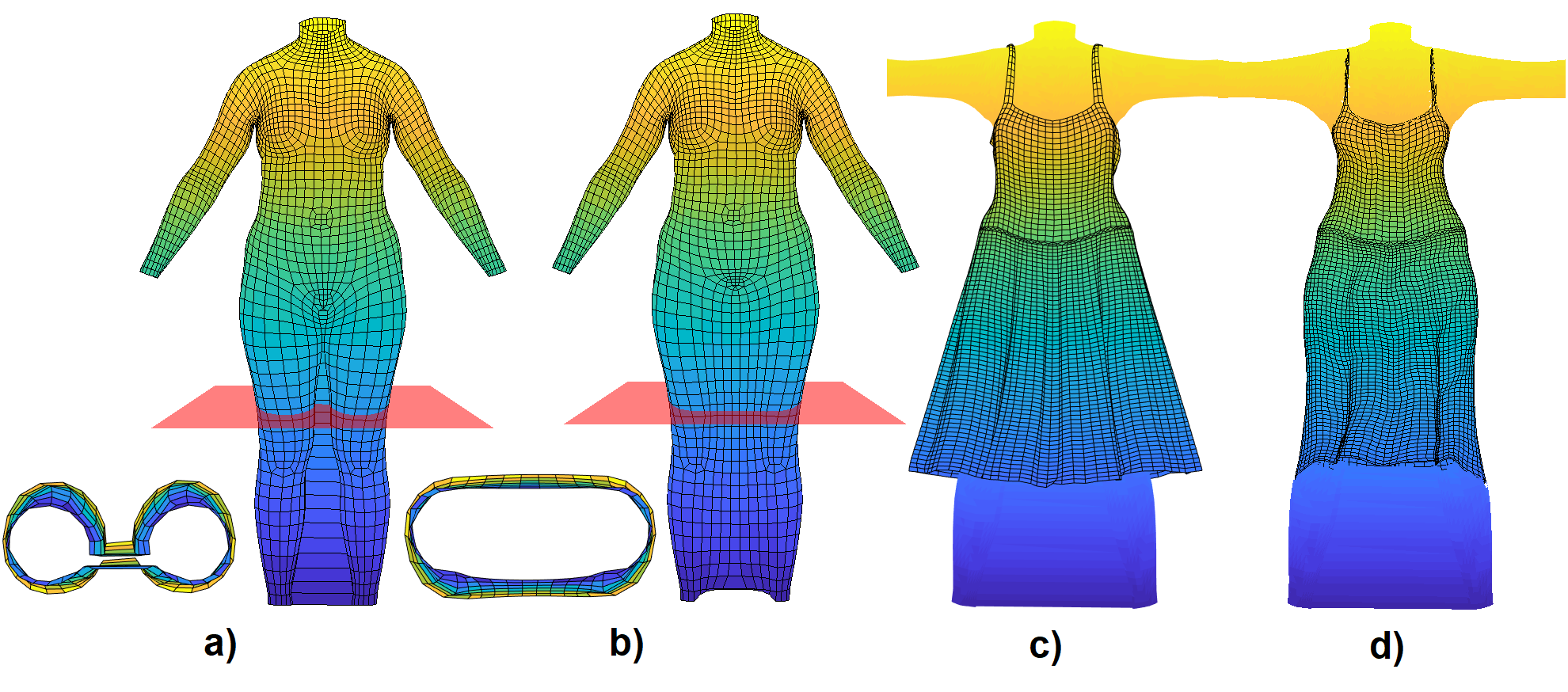}
    \caption{Dual topology and registration. a) New additional proposed topology, where inner legs are connected. This topology is used for graph convolutions as well. b) Result of Laplacian smoothing of inner leg vertices. It is used only for skirt/dress registration. We show top view of meshes around an imaginary red cutting plane. c) Garment in rest pose. d) Garment registered to body model.}
    \label{fig:registration}
\end{figure}

\subsection{Data Pre-processing}
\label{sec:data}

In order to match among garment and body, we apply non-rigid ICP \cite{amberg2007optimal}. Registration is performed once per sequence in rest pose. Due to SMPL low vertex resolution, garment details could be lost. For this reason we subdivide the mesh (and corresponding SMPL model parameters). Head, hands and feet are not used to find correspondences and 
removing them halves input dimensionality. This yields a final mesh with $N=14475$ vertices. Finally, note that skirt-like garments do not follow the same topology as SMPL mesh. For this task we introduce a novel topology explained on the subsection below. An example of the registration is shown in Fig. \ref{fig:registration}. Finally, body to cloth correspondences and garment mask are extracted by nearest neighbor matching.

\subsection{SMPL-Skirt Topology}
From SMPL body mesh, a ‘column’ of inner faces of each leg is removed and a new set of faces is created by connecting vertices from both legs, see Fig.4a. New faces are highly stretched, producing noisy garment registrations if used as is, NR-ICP yields optimal results for homogeneous meshes (in garment domain). Because of this, we apply an iterative Laplacian smoothing to vertices belonging to the inner parts of each leg, see Fig.4b for the result. This process is repeated before registration with the corresponding shape of the subject in the sequence in T-pose. This gives a matching between garment and body vertices to compute offsets. For encoding garments as offsets we use body mesh without smoothing, as this process will misbehave for posed bodies. Finally, for graph convolutions, we use the Laplacian matrix corresponding to this new topology for garments of type Dress and Skirt. This ensures that vertex deep features are aggregated with the correct neighbourhood. Afterwards, we transfer body topology to the predicted garment, and it is therefore crucial to use the correct topology for each garment type.

\subsection{Network}
\label{sec:gcvae}

\begin{figure*}[ht!]
    \centering
    \includegraphics[width=.95\textwidth,height=4cm]{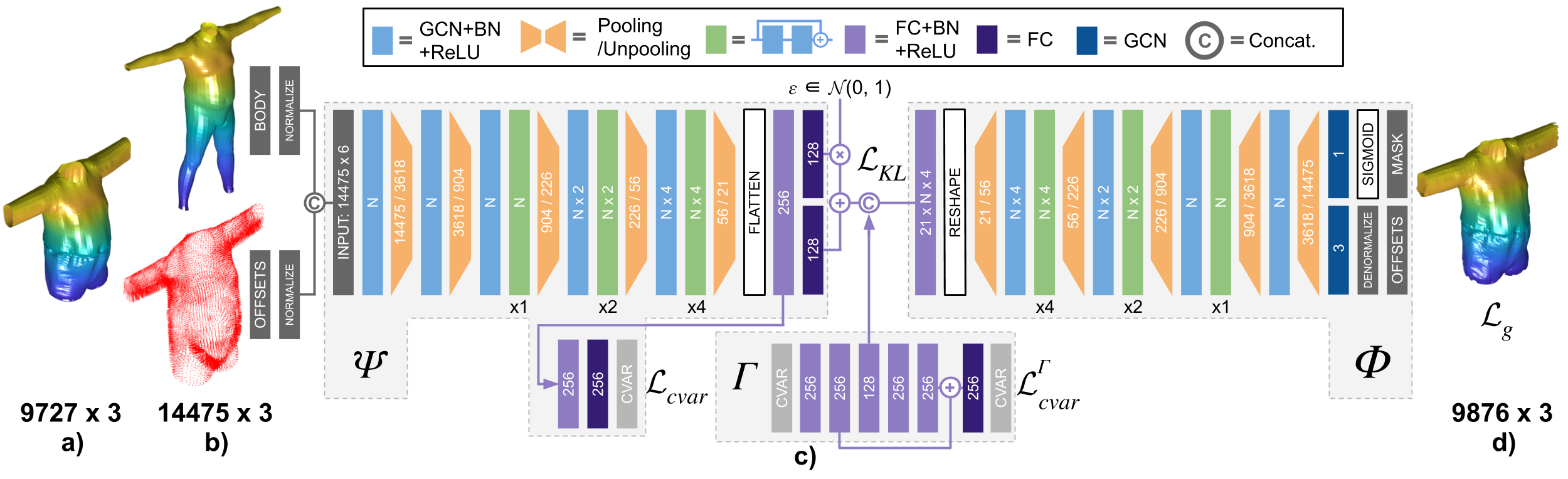}
    \caption{Model pipeline. a) Input garment b) body and offsets w.r.t. body (Sec. \ref{sec:data}). Model input is the concatenation of body and offsets. c) Network architecture. Conditional variables (CVAR) are processed by an AutoEncoder. To improve latent space factorization, CVAR are also regressed from the first encoder FC layer. Decoder outputs are offsets and mask. d) Reconstruction of the garment by adding offsets to body and removing body vertices according to mask. We set $N$ as 128.}
    \label{fig:pipeline}
\end{figure*}

As shown in Fig. \ref{fig:pipeline}, our network is based on a VAE generative model. The goal is to learn a meaningful latent space associated to the garments of any type, shape or with wrinkles which is used to generate realistic draped garments. Garment type and shape are associated to the static state of the garment while wrinkles belong to the dynamics of the garments. Here, we disentangle the latent space between statics and dynamics of the garments, and refer to learnt latent codes as garment code ($z_s\in\mathbb{R}^{128}$) and wrinkle code ($z_d\in\mathbb{R}^{128}$), respectively. To do so, we build two separate networks, one trained on static garments (so called SVAE) and one trained on dynamic garments (so called DVAE). To factorize the latent space from irrelevant parameters to the garment type and shape, we condition SVAE on body shape ($\beta\in\mathbb{R}^{11}$)\footnote{We include gender as an additional dimension to the shape parameters.} and garment tightness ($\gamma\in\mathbb{R}^2$). Likewise, DVAE is conditioned on $\beta$, $\gamma$, body pose ($\theta\in\mathbb{R}^{f\times72}$) and $z_s$, where $f$ is the number of frames in a temporal sequence. Let $cvar_s$ and $cvar_d$ be the stacking of conditioning variables of SVAE and DVAE in a single vector. It is worth noting that $\theta$ is constant in SVAE so that we do not include it in $cvar_s$. We implement graph convolutions as in \cite{bronstein2017geometric,defferrard2016convolutional,ma2019dressing,niepert2016learning,wu2019comprehensive,yuan2019mesh}. We also include skip connections throughout the whole network.

\begin{figure}[!t]
    \centering
    \includegraphics[width=.6\textwidth,height=4cm]{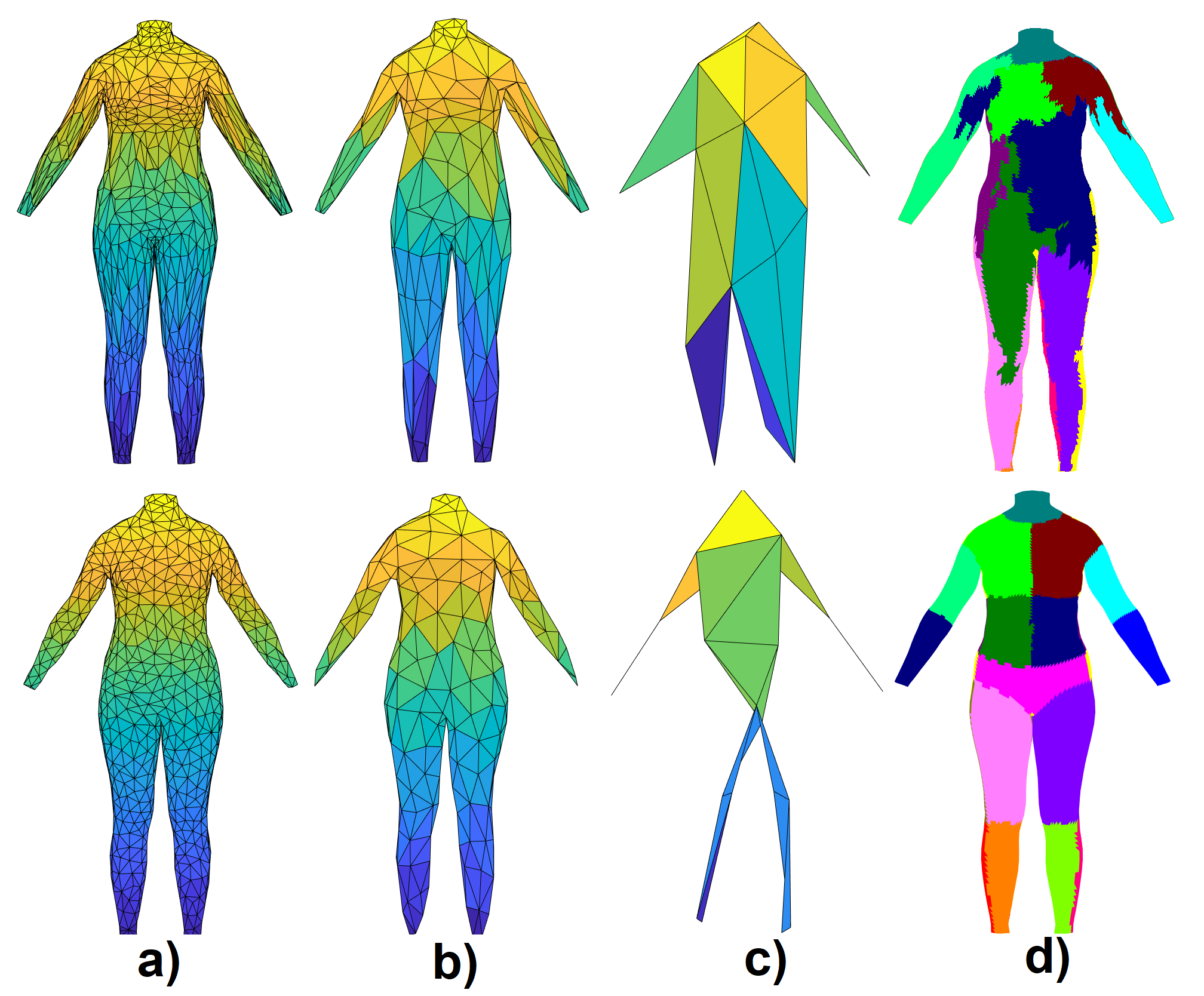}
    \caption{Mesh hierarchy for pooling. Upper: default~\cite{garland1997surface}. Lower: proposed. a), b) and c) depict the mesh hierarchy used for graph pooling through the model. Observe the difference on spatial distribution at a) and b). c) shows how lowest pooling is more meaningful regarding the segments (one vertex per segment). d) is the visualization of correspondences (receptive field) between highest and lowest hierarchy levels. The proposed pooling yields more meaningful pooling receptive fields w.r.t. body parts.}
    \label{fig:mesh_hierarchy}
\end{figure}

\textbf{Architecture.} Let $X_s \in \mathbb{R}^{V_T\times 3}$ and $X_d \in \mathbb{R}^{V_T\times 3}$ be offsets computed on static and dynamic samples, respectively. From now on we use subscript $s$ and $d$ for static and dynamic variables and discard them for general cases. SVAE and DVAE have a similar structure with three main modules: encoder $\{cvar^z, z\}=\Psi(\bar{X}, \bar{T})$, conditioning $\{cvar, cvar^z\}=\Gamma(cvar)$ and decoder $\{\bar{X}, M\}=\Phi(z, cvar^z)$, where $M\in\mathbb{R}^{V_T\times1}$ is the garment mask. Conditioning network $\Gamma$ is an autoencoder with one skip connection and $cvar^z$ is its middle layer features. The goal of this network is to provide a trade-off between $cvar$ and $z$. The architecture details are shown in Fig. \ref{fig:pipeline}. Note that all GCN layer features (except first and last layers) are doubled in DVAE vs. SVAE. We refer the reader to the supplementary material for more details on the network architecture.

\textbf{Pooling.} We resort to a mesh simplification algorithm \cite{garland1997surface} to create a hierarchy of meshes with decreasing detail in order to implement the pooling operator. We follow \cite{yuan2019mesh} to have vertices uniformly distributed in the graph coarsening. However, this approach does not guarantee a uniform or meaningful receptive field on a high resolution mesh. To achieve a homogeneous distribution of correspondences throughout the body between pooling layers, we define a segmentation (Fig. \ref{fig:mesh_hierarchy}(d)) and forbid the algorithm from contracting edges connecting vertices of different segments. Segmentation contains $21$ segments and it is designed such that regions of the body with highest offset variability have smaller segments. Thus, more capacity of the network is available to model those parts. See Fig. \ref{fig:mesh_hierarchy}. Our mesh hierarchy is formed by $6$ different levels. The dimensionality of those meshes is: $14475\rightarrow 3618\rightarrow 904\rightarrow 226\rightarrow 56\rightarrow 21$, leaving a single node for each segment on the last pooling layer. We use max-pooling in the proposed hierarchy. For unpooling, features are copied to all corresponding vertices of the immediate higher mesh.

\textbf{Loss.} We train conditioning network $\Gamma$ independently using $L1$ loss and freeze its weights while training VAE. S/DVAE loss is a combination of a garment related term, a $cvar$ term and KL-divergence:
\begin{equation}
    \mathcal{L} = \mathcal{L}_{g} + \mathcal{L}_{cvar} + \lambda_{KL} D_{KL}(q(z|X,cvar)||p(z|cvar)),
\end{equation}
Garment related term handles offsets, mask (if available), smoothness and collisions:
\begin{equation}
    \mathcal{L}_{g} = \mathcal{L}_{o} + \lambda_{n}\mathcal{L}_{n} + \lambda_{m}\mathcal{L}_{m} + \lambda_{c}\mathcal{L}_{c},
\end{equation}
where $\mathcal{L}_o$ is an L1-norm applied to output offsets. $\mathcal{L}_n$ is the smoothness term based on L1-norm on normals. We found that regular Laplacian loss ensures smoothness at the cost of losing high frequency geometric details, while a normal loss makes output geometry consistent w.r.t. the input. $\mathcal{L}_m$ consists on L1-norm on mask. Finally, $\mathcal{L}_c$ is the collision loss. Given that garments are represented as offsets, we design this loss as:
\begin{equation}
    \mathcal{L}_{c} = max(0, -o\cdot V_{N}),
\end{equation}
where $o$ are the output offsets and $V_N$ are the body normals at the corresponding vertices, this penalizes offsets that go within the body. $\mathcal{L}_{cvar}$ is $L1$ loss on encoder $cvar^z$ regressor.

\section{Experiments}
\label{sec:experiments}
    First, we detail the metrics chosen to analyze the results.

\textbf{Surface.} Given that input and prediction have the same dimensionality and order, we use standard euclidean norm (in mm.).

\textbf{Normals.} Measure of surface quality. We compute normals error based on mesh face normals by their angle difference (in radians) to ground truth normals.

\textbf{Mask}. Garment mask is evaluated by the intersection over union (IoU).

\textbf{KL Loss}. We use KL loss as a measure of quality of latent code factorization and meaningfulness of the latent space.

\subsection{Ablation Study}

\begin{table*}[!t]
\centering\scriptsize\setlength\tabcolsep{1.5pt}
\begin{subtable}{.5\textwidth}
\centering

\begin{tabular}{|l||c|c|c|c|c|}
\hline
 & Surface & Normals & Mask & KL loss \\ \hline \hline
All & 14.3 & 1.04 & 0.9518 & 0.9820 \\ \hline
No normals & 22.8 & 1.07 & 0.9472 & 0.5966 \\ \hline
No mask & 92.7 & 1.19 & - & 0.8799 \\ \hline
No collision & 14.7 & 1.02 & 0.9522 & 0.9414 \\ \hline
No CVAR & 14.8 & 1.02 & 0.9520 & 1.1009 \\ \hline
Default pooling & 14.9 & 1.03 & 0.9390 & 0.7623 \\ \hline
\end{tabular}
\caption{}
\label{tab:ablation-all}
\end{subtable}
\begin{subtable}{.5\textwidth}
\centering

\begin{tabular}{|l||c|c|c|c|}
\hline
 & Surface & Normals & Mask & KL loss \\ \hline \hline
Top & 11.9 & 1.20 & 0.9035 & 0.9536 \\ \hline
T-shirt & 15.5 & 1.21 & 0.9565 & 1.1701 \\ \hline
Trousers & 10.9 & 0.84 & 0.9475 & 0.9008 \\ \hline
Skirt & 21.4 & 0.79 & 0.9520 & 1.0255 \\ \hline
Jumpsuit & 13.3 & 1.07 & 0.9637 & 0.8788 \\ \hline
Dress & 16.7 & 1.06 & 0.9662 & 0.9995 \\ \hline
\end{tabular}
\caption{}
\label{tab:ablation-per-garment}
\end{subtable}
{\normalsize
\subcaption{Table 2: (a) Ablation results on the static dataset for all clothes. (b) Ablation results (full model) on the static dataset for each cloth category. Surface and normal errors are shown in mm and radians, respectively.}
}
\end{table*}

\begin{table*}[!t]\centering\small
\begin{tabular}{|l||c|c|c|c|c|c|c|}
\hline
\# frames & Top & T-shirt & Trousers & Skirt & Jumpsuit & Dress & Avg. \\ \hline \hline
1 & 21.8/1.24 & 28.8/1.29 & 20.7/0.89 & 37.6/0.92 & 28.2/1.15 & 35.5/1.13 & 29.0/1.10 \\ \hline
4 & 20.1/1.23 & 28.0/1.28 & 18.5/0.86 & 33.2/0.89 & 26.1/1.09 & 32.2/1.11 & 26.1/1.08 \\ \hline
\end{tabular}
    \caption{Ablation results (full model) on the dynamic dataset conditioning on different number of frames. Left: surface error (mm) / Right: normals error (radians).}
\label{tab:ablation-dynamic}
\end{table*}

We trained SVAE on an additional dataset of static samples (in rest pose) with 30K samples. 20\% of the data is kept for evaluation and the rest for training. The results are shown in Tab. \ref{tab:ablation-all} and \ref{tab:ablation-per-garment}.

\textbf{Normals.} Looking at the second row of Tab.\ref{tab:ablation-all} we observe that enforcing a reconstruction consistent with normals significantly reduces surface error and, as expected normals error. However, including normals has a negative impact on KL loss comparing to first row.

\textbf{Mask.} As seen in third row of Tab. \ref{tab:ablation-all}, both, surface and normals error are significantly higher without mask prediction (comparing to first row).

\textbf{Collision.} Fourth row of Tab. \ref{tab:ablation-all} shows how collision loss helps to improve vertex location by pushing collided vertices to their correct position. On the other hand, it is observable a non-significant increase on other losses. 

\textbf{CVARs.} As explained in Sec.\ref{sec:gcvae}, conditional variables are regressed from the first FC layer of the encoder to improve latent space factorization. On fifth row of Tab. \ref{tab:ablation-all} we can see that, while surface or normals error have no significant differences, KL loss improves.

\textbf{Pooling.} On Sec.\ref{sec:gcvae} we discussed different approaches for tackling the pooling on a graph neural network. To do this, we built a mesh hierarchy. We compared default mesh simplification algorithm versus our proposed modification. Results are shown in the last row of Tab. \ref{tab:ablation-all}. While improvement on surface and normals errors is marginal, this new pooling benefits mask prediction.

\textbf{Per Garment Category Error.} Results per garment are shown in Tab. \ref{tab:ablation-per-garment}. Skirts present the highest surface error, as its vertices are further away from the body compared to other garments. Following this reasoning, we find trousers having the less surface error. If we look at normals error, we find an opposite behaviour for skirts, as their geometry is the simplest one. On the other hand we see that upper-body garments present more complex geometries, and therefore, higher normals error. Looking at mask error, we see that garments that cover most of the body have the lowest error. This is due to IoU metric nature, the lower the number of points, the more impact shall have each wrong prediction. Finally, looking at KL loss, we observe the model has difficulties to obtain meaningful spaces for T-shirts. As explained on Sec.\ref{subsec:garment_generation}, T-shirts category includes open shirts as well, which highly increases class variability. We also see that trousers and jumpsuits have the lowest KL loss.

\begin{figure}[!t]
    \centering
    \includegraphics[width=.85\textwidth,height=6cm]{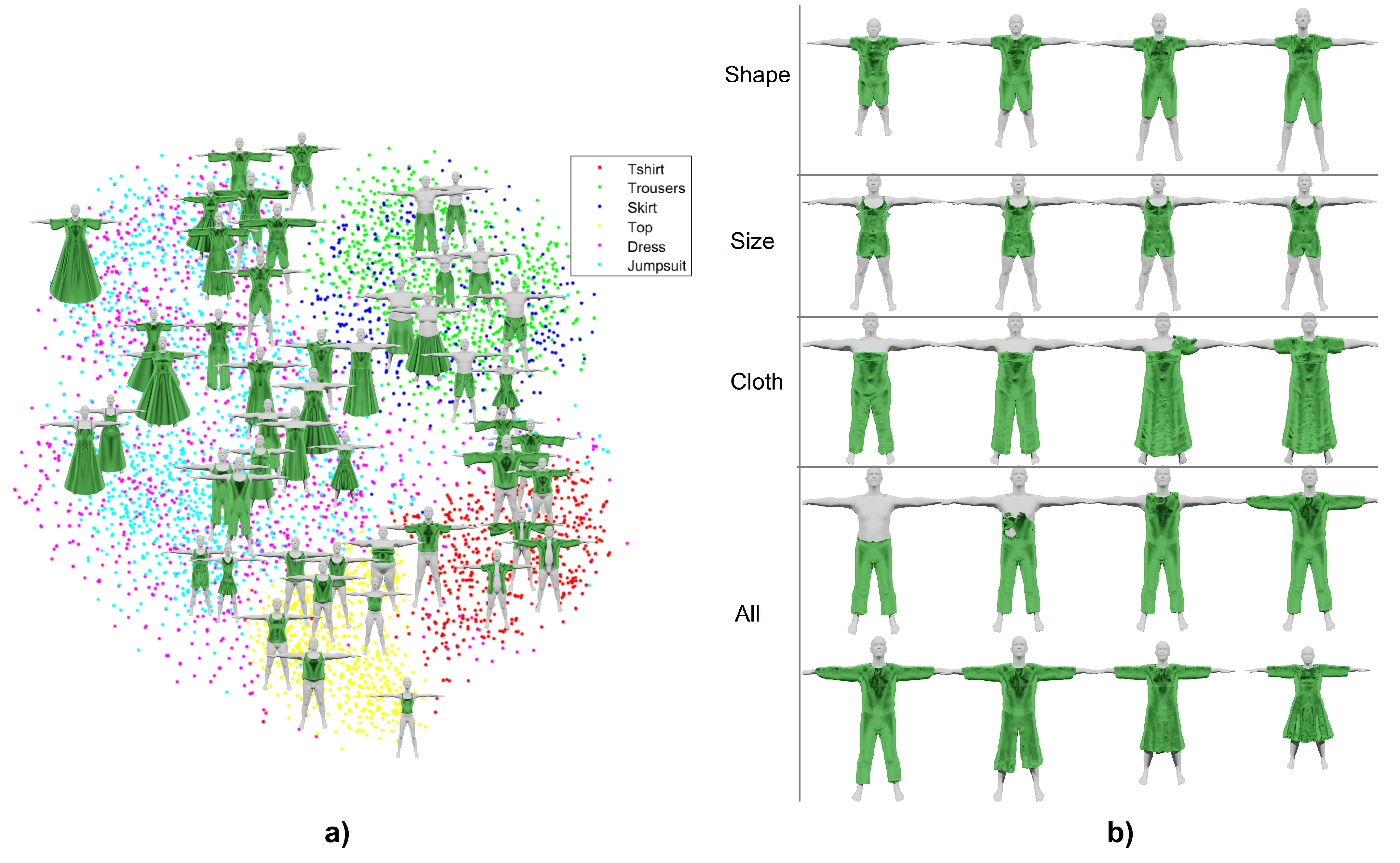}    
    \caption{a) Visualization of the learned latent space for static samples using t-SNE algorithm. b) Transitions of static samples. First three rows: conditioning on shape, tightness or cloth while the rest are fixed. Last two rows: transition of all variables. Variables are linearly graduated.}
    \label{fig:tsne}
    \end{figure}

\textbf{Learned Latent Space.} In Fig. \ref{fig:tsne}a, we show distribution of 5K random static samples computed by t-SNE algorithm. As one can see, the proposed GCVAE network can group garments in a meaningful space. Interestingly, dress and jumpsuit that share more vertices also share the same latent space. Additionally, we show garment transitions in this space in Fig. \ref{fig:tsne}b. One can see how garments transit between two different topologies (3rd row) or among different genders and shapes (4th row).

\begin{figure}[!t]
    \centering
    \includegraphics[width=0.95\textwidth]{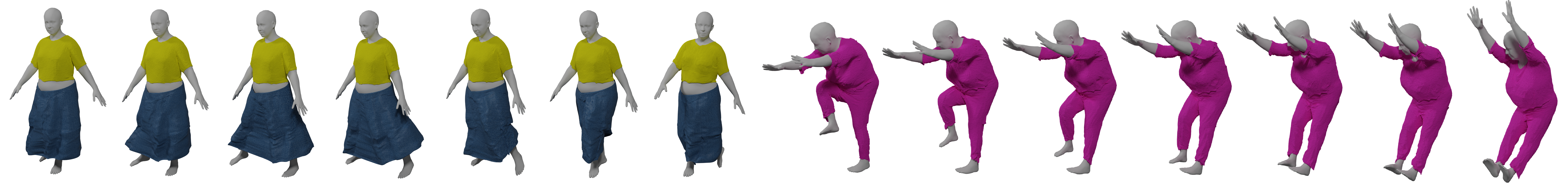}
    \caption{Garment reconstruction for sequences. Note that the model has not been trained to keep temporal consistency.}
    \label{fig:qual_dyn}
\end{figure}

We study DVAE model in Tab. \ref{tab:ablation-dynamic}. We condition DVAE on pose for a single frame vs. four frames. Four frames are selected every 3 frames, resulting in a 12-frame clip. Training the model on a sequence of frames leads to better results in all garment categories (3mm improvement in average). This is while we do not include any temporal information in the encoder nor any specific sequence prediction loss. DVAE qualitative results for single frames and sequences are shown in Fig.\ref{fig:title}(right) and Fig.\ref{fig:qual_dyn}, respectively.

\section{Conclusions}
\label{sec:conclusion}
We presented CLOTH3D, the first large scale synthetic dataset of 3D clothed humans. It has a large data variability in terms of body shape and pose, garment type, topology, shape, tightness and fabric. Generated garments also show complex dynamics, providing with a challenging corpus for 3D garment generation. We developed a baseline method using a graph convolutional network trained as a variational autoencoder, and proposed a new pooling grid. Evaluation of the proposed GCVAE on CLOTH3D showed realistic garment generation.

\textbf{Acknowledgments.} This work is partially supported by ICREA under the ICREA Academia programme, and by the Spanish project PID2019-105093GB-I00 (MINECO / FEDER, UE) and CERCA Programme / Generalitat de Catalunya.

\pagebreak

{\small
\bibliographystyle{splncs04.bst}
\bibliography{egbib.bib}
}

\chapter*{CLOTH3D: Supplementary Material}

This part of the document provides additional details about CLOTH3D dataset and ablation analyses. First, Sec.\ref{sec:cloth3d_details} provides further descriptions on the data generation process. Secondly, Sec.\ref{sec: var} describes the variability and size of the dataset. Then, Sec.\ref{sec:technical_details} gives specific details on data format. Along with this document a video is submitted. The content of the supplementary video is described in Sec.\ref{sec:video}. Later, in Sec.\ref{sec:method} we provide more details about the implemented Graph CNN (Sec. \ref{sec: gcn} and \ref{sec: arch}), further discussions about wrinkle factorization (Sec. \ref{sec: factorization}), qualitative generated results from the learnt latent space (Sec. \ref{sec: results}), and analysis on the DVAE error distribution (Sec. \ref{sec:dyn}). Finally, Sec. \ref{sec:applications} describes theoretical potential applications of CLOTH3D dataset.

\section{CLOTH3D}
\subsection{Dataset generation details}
\label{sec:cloth3d_details}

In this section we explain in more detail relevant aspects of the dataset generation process. First, we introduce a summary of the generation steps. Then, we start on the human sequence generation (Sec.3.1 on main paper), with the specifics on self-collision solving. Later, we move to garment generator (Sec.3.2 on main paper) shaping step, where a formal description is given, plus a clarification on how to keep tightness semantically meaningful w.r.t. SMPL shape displacements. Finally, we describe an important pre-requisite for simulation (Sec.3.3 on main paper) that allows the simulation of any sequence from rest pose.

\textbf{Generation algorithm.} Here we briefly summarize the steps performed by our generator:

\setlist[enumerate]{label*=\arabic*.}
\begin{enumerate}[noitemsep]
    \item HUMAN
    \begin{enumerate}[noitemsep]
        \item Pick SMPL parameters
        \item Compute SMPL body sequence
        \item Solve self-collisions
    \end{enumerate}
    \item OUTFIT
    \begin{enumerate}[noitemsep]
        \item Pick template garments
        \item Shape sleeves/legs/skirt
        \item Cut
        \item Resize
        \item Sew garments into jumpsuit/dress (optional)
    \end{enumerate}
    \item SIMULATION
    \begin{enumerate}[noitemsep]
       \item Fabric settings
        \item Body shape transition (from $\beta+\gamma$ to $\beta$)
        \item Pose transition
        \item Simulate sequence
     \end{enumerate}
\end{enumerate}

\textbf{Self-collision.} As explained on Sec.3.1 of the main paper, SMPL can present self-collisions, specially on samples with challenging shapes. This makes cloth simulation difficult for many combinations of shape and pose. In order to overcome this and drastically increase the amount of valid samples, we propose a simple, fast self-collision solving step. Through visual inspection, we identified problematic body regions (armpit, crotch, ...) on which to detect and solve collisions. Using SMPL segmentation, we separate vertices belonging to different body parts. Collisions appear as intersection of pairs of segments. For each of these pairs, we test edges of a segment vs faces of the other, and viceversa. Since we identified problematic regions, the number of edge-vs-face tests is significantly reduced. This yields a set of intersection points to which we approximate a plane. Then, each collided body vertex is moved to the corresponding side of this plane based on segment index. We leave a separation of $4$mm as for simulation, a $2$mm margin is used for cloth-body collisions.

\textbf{Resizing.} In Sec.3.2 of the main paper, we explained how we resized garments with SMPL shape parameters plus an offset to represent garment tightness. While first and second shape parameters represent overall size and fatness respectively, the sign of the first parameter has opposite meaning for male and female. To take this into account and so that tightness remains semantically consistent, the sign of the first offset is $(-1)^{g+1}$, where $g$ is gender ($0 =$ female and $1 =$ male). This means a positive tightness shall produce smaller garments.

\textbf{Body transition.} Outfit generation process yields garments on rest pose resized to SMPL shape plus tightness. We need a body transition from this state to the state of the initial frame of the sequence for a correct simulation. To ensure no body-to-cloth penetration is present due to resizing to a different shape, we generate a few frames of transition where the body shape changes from $\beta+\gamma$ (shape+tightness) to $\beta$. Finally, more frames are devoted to a transition from rest pose to the initial pose of the sequence split. Pose transition is computed based on quaternions for a smooth posing.

\subsection{Variability and Size}
\label{sec: var}

Each template garment is shaped with linear deformations, similar to SMPL shaping, where coefficients are uniformly sampled to yield a balanced distribution. Garments are cut at uniformly sampled lengths along limbs and waist/torso. With this we potentially obtain all possible combinations of sleeve length and t-shirt length, or leg length and waist height. Furthermore, upper-body garments have specifically designed cuts that change shape and topology, also uniformly sampled. Finally, on resizing stage, a tightness two-dimensional factor is uniformly sampled from $[-1.5, 0.5]$ (biased to loose garments for more complex dynamics and wrinkles), further increasing garment variability. All these randomly generated properties, once combined, almost guarantee that an outfit will never appear twice in the dataset. Afterwards, different simulation properties will also ensure cloth shall behave differently. Each pose sequence from source data is downgraded from $60$fps to $30$fps and split into $300$ frames subsequences, each of them simulated once with a different outfit, thus totalling around $2$M frames.

\subsection{Data format}
\label{sec:technical_details}

Each sample is a $600$-frames split of the original sequence on source data, downgraded from $60$fps to $30$fps, totalling $300$ frames for each split. Note that frames refer to time instants, not images. The name of the sample contains the name of the original sequence and the number of the split (e.g.: '01$\_$01$\_$s0', sequence is '01$\_$01' and split is 's0'). As the number of frames of original sequences is not a multiple of $300$, some splits will have smaller length. Each sample has static and dynamic garment information. \textbf{Static} information is the outfit fitted to rest pose for the corresponding body shape. Static garments are represented by OBJ files, which include vertices on rest position and topology data. The \textbf{dynamic} information contains garment animation data. 3D animation data is usually stored as PC2 (Point Cache 2) files. We propose the PC16 format, a PC2 conversion from 32-bit floats to 16-bit, halving dataset space requirements. Precision loss on this conversion is none to minimal within the range $[-1,1]$ ($\leq2^{-11}$) and insignificant in $[-2, 2]$ ($\leq2^{-10}$). By storing garment vertex positions relative to SMPL body root joint, we ensure values will always be in range $[-2, 2]$. Sample metadata contains SMPL parameters, garment names and their fabrics. To ease the fitting process, rest pose is redefined such that legs are slightly open, due to the high geometric complexity of this region. Average number of vertices per outfit is around $20$K, which implies an average size of $35-40$MB per sequence.

\begin{figure*}
    \centering
    \includegraphics[width=0.85\textwidth]{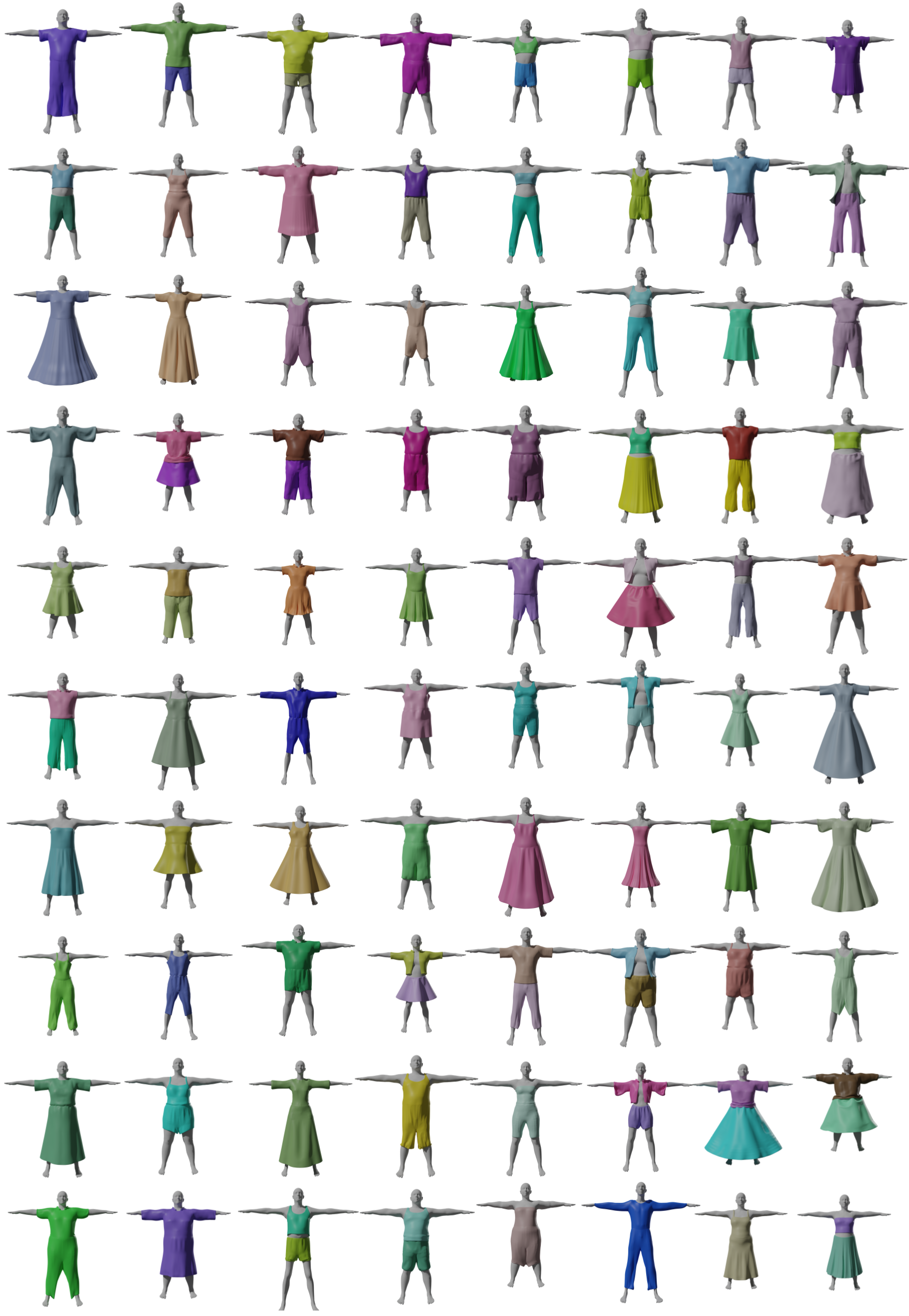}
    \caption{CLOTH3D static samples.}
    \label{fig:static}
\end{figure*}

\begin{figure*}
    \centering
    \includegraphics[width=.84\textwidth]{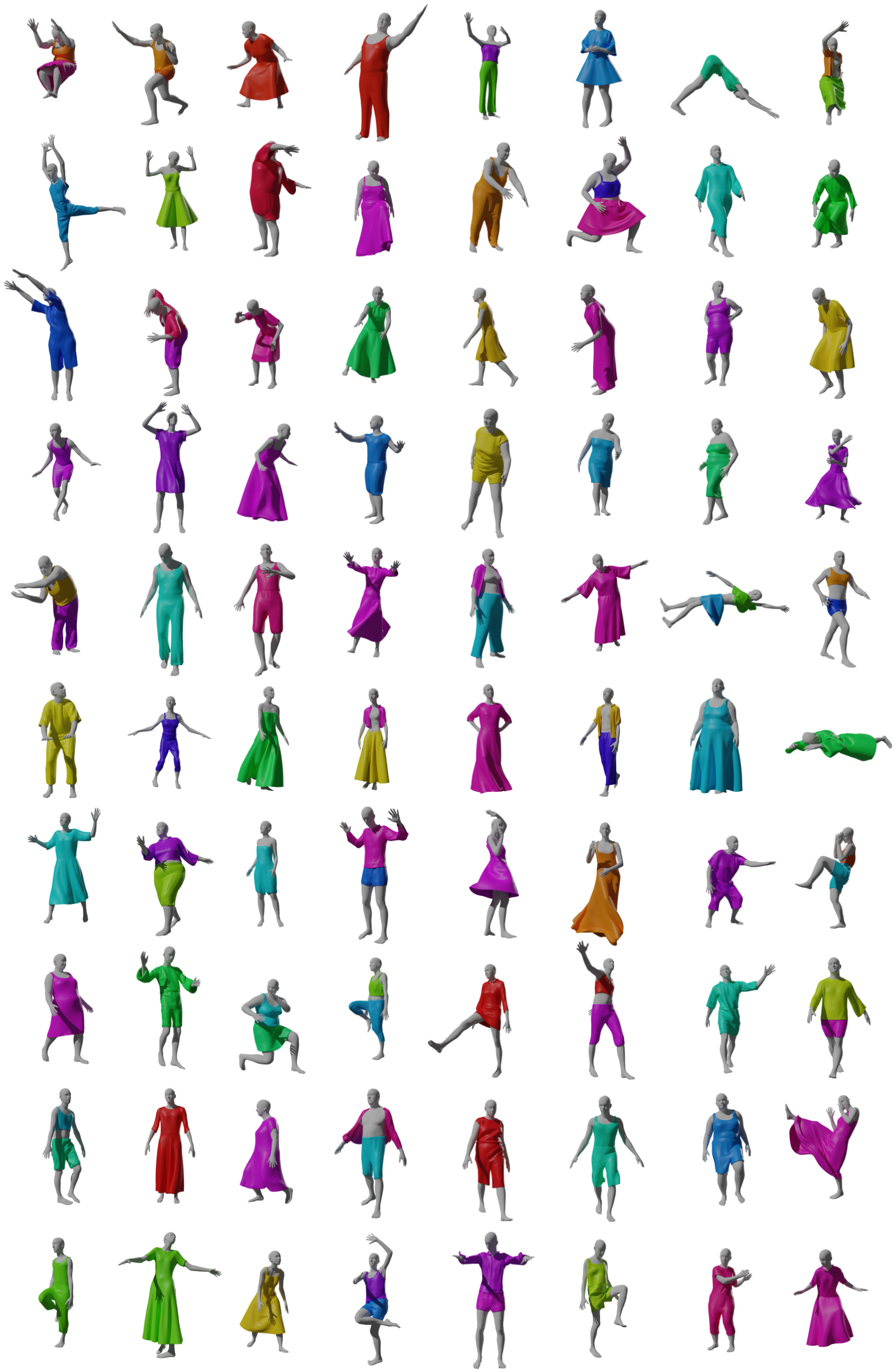}
    \caption{CLOTH3D dynamic samples.}
    \label{fig:dynamic}
\end{figure*}

\begin{figure*}
    \centering
    \includegraphics[width=.85\textwidth]{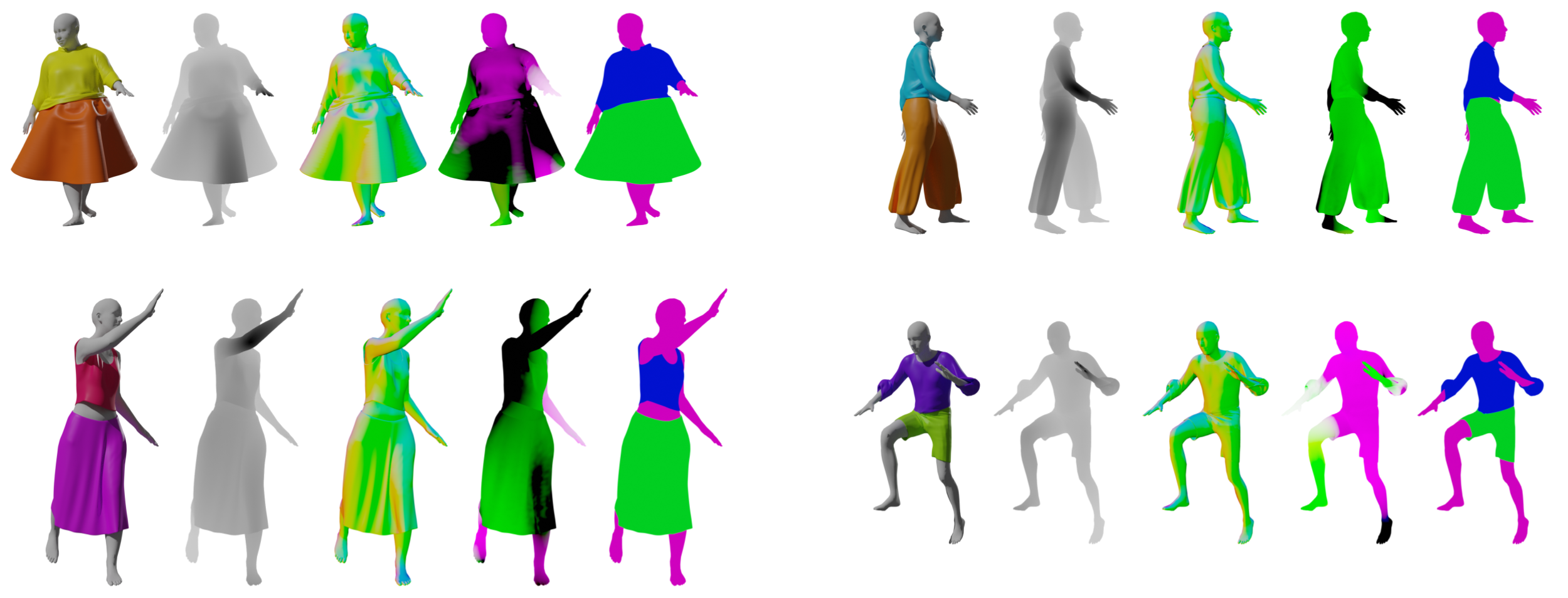}
    \caption{CLOTH3D data representations. From left to right: RGB, depth, normals, velocities, and segmentation masks.}
    \label{fig:modalities}
\end{figure*}

Fig. \ref{fig:static} shows random static samples. Fig. \ref{fig:dynamic} shows random frames of sequences. Finally, Fig. \ref{fig:modalities} shows random samples of different sequences with different representation modalities that can be obtained from CLOTH3D data: depth maps, surface normals, 3D velocities and segmentation masks.


\section{Video}
\label{sec:video}

We provide a video supplementary material which visualizes a few samples of CLOTH3D. We further show different data representations provided at the dataset for those samples. The top-left image of the video is a 3D visualization as RGB data. We chose high quality lighting and shaders to make more evident the geometric details of the 3D models. On top-right part of the image, we show a rendering of depth information w.r.t. camera. Depth maps are normalized per frame. Then, the bottom-left image shows an encoding of 3D surface normals as RGB colors. Since normals are already within range $[0, 1]$, only a mapping to RGB is performed. This means each color channel represents a 3D coordinate of the normal vectors. Finally, the bottom-right image renders 3D velocities. As for normals, they are encoded as RGB data, and, as depth, it is normalized per frame. Due to normalization, frames with low motion appear noisy.

\section{Dressed human generation}
\label{sec:method}

As we explained in Sec.4 in the main paper, we apply a variational autoencoder with graph convolutional neural (GCN) blocks to learn some latent codes. Later in test time we sample from the latent space to generate dressed humans. We condition the model on the garment type, tightness, body shape and pose. Conditioning on the garment type can be done by a one-hot encoding of the garment category (e.g. T-shirt, dress, etc.). Although we have fixed garment categories in the dataset, their variability in size, shape or type allows us to encode garments into a continuous space, which is more meaningful than fixed categories to be conditioned on. Therefore we train our model on static garments (in the rest pose) to learn the garment type code. We briefly remind the notations from the main paper for the easiness of reading. Here, we provide additional details of the implemented GCN, Sec.\ref{sec: gcn}. Then, Sec.\ref{sec: factorization} explains our wrinkles factorization procedure. Finally Sec.\ref{sec: results} shows some generated results.

We train SVAE on static garments and DVAE on dynamic garments and refer to learnt latent codes as garment code ($z_s\in\mathbb{R}^{128}$) and wrinkle code ($z_d\in\mathbb{R}^{128}$), respectively. Here, we provide DVAE details. DVAE has three main modules: encoder $\{cvar^z, z\}=\Psi(\bar{X},\bar{T};\tau^{\Psi})$, conditioning $\{cvar, cvar^z\}=\Gamma(cvar;\tau^{\Gamma})$ and decoder $\{\bar{X}, M\}=\Phi(z,cvar^z;\tau^{\Phi})$. $X \in \mathbb{R}^{V_T\times 3}$ are the offsets computed on dynamic samples, $\bar{X}$ is a normalization of $X$, $T\in \mathbb{R}^{V_T\times 3}$ and $\bar{T}$ are body vertices and its normalization, $cvar$ is the stacking of conditioning variables (body shape $\beta$, garment tightness $\gamma$, body pose $\theta$ and garment code $z_s$), $cvar^z$ is the middle layer features of autoencoder network $\Gamma$, $M\in\mathbb{R}^{V_T\times1}$ is the garment mask and $\tau^{.}$ are networks weights. Finally, $C\in \mathbb{R}^{V_T\times3}$ are the garment vertices after registration on top of $T$.

\subsection{Graph convolutions}
\label{sec: gcn}
Sec.4.2 of the main paper describes the model architecture. Here we focus on the formulation for graph convolutions we implemented. Our network input data is defined by $V_T$ and topology. This allows to use graph convolutional filters to learn features. Following the definition of spectral graph convolution, filtering is computed as:
\begin{equation}
    y = g_{\omega}(\mathbf{L})x = \sum_{i=0}^K\omega_i \mathbf{T}_i(\hat{\mathbf{L}})x,
\end{equation}
where $\omega_i$ are the learnable filters, $\mathbf{L}$ is the normalized Laplacian matrix, $\hat{\mathbf{L}} = 2\mathbf{L}/\lambda_{max} - \mathbf{I}$, and $\mathbf{T}_i(\hat{\mathbf{L}})$ is the $i$-th Chebyshev polynomial order. Given an input graph with $F_{in}$ features for each node, the described convolution will return that same graph with a different set of features $F_{out}$. Chebysev polynomial order defines the size of the receptive field $K$, meaning feature filtering aggregates the $K$-ring neighbourhood for each node. In our case, we keep an small neighborhood with $K=1$. Therefore to have a high receptive field we build a deep network. Each convolutional and pooling layer further combines node features with higher $K$-ring neighbours. We also include skip connections throughout the whole network. This leads to an effective information passing helping to learn fine-grain garment details.

\subsection{Architecture details and justification}
\label{sec: arch}

As a standard operation in CVAE, conditional variables should be fed into the encoder. However, these variables ($cvar$) are not balanced with offsets ($X$) in terms of size and scale. $cvar$ can be partially decoded to body vertices $\bar{T}$ and offsets $\bar{X}$. Therefore, we concatenate $\bar{T}$ to $\bar{X}$ and feed it to the network $\Psi$. To better factor out $cvar$ from latent code $z$ we include an additional MLP branch at the end of the encoder and before sampling layer. The goal of this branch is to regress $cvar$ during training. It also helps to have a more stable training. Regularizing the encoder by regressing $cvar$ has a limitation when the dimensionality of $cvar$ is high (e.g. $cvar_d$), that is, optimization can stick to local minima. Therefore, we regress $cvar^z$ instead of $cvar$. Finally, decoder generates normalized offsets $\bar{X}$ and garment mask $M$ in two branches at the last layer. Note that DVAE does not have branch $M$ in the decoder. Note that we train our GCN with dual topology to handle skirt-like garments vs. the rest which has not been done before in 3D garment reconstruction.

\begin{figure*}[!ht]
    \centering
    \begin{subfigure}{0.48\textwidth}
        \includegraphics[width=\textwidth]{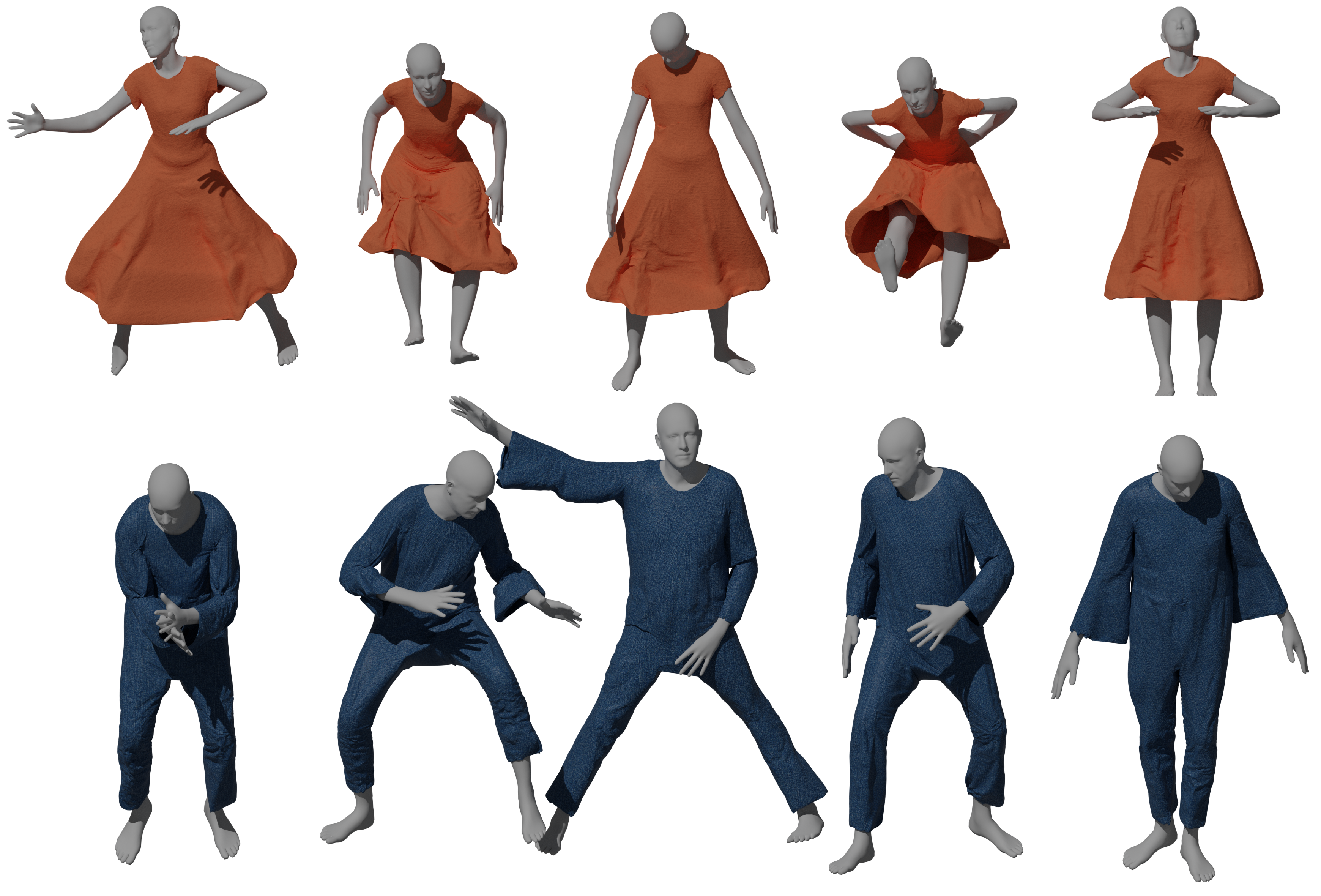}
        \caption{Pose variation.}
        \label{fig:trans_pose}
    \end{subfigure}
    \begin{subfigure}{0.48\textwidth}
        \includegraphics[width=\textwidth]{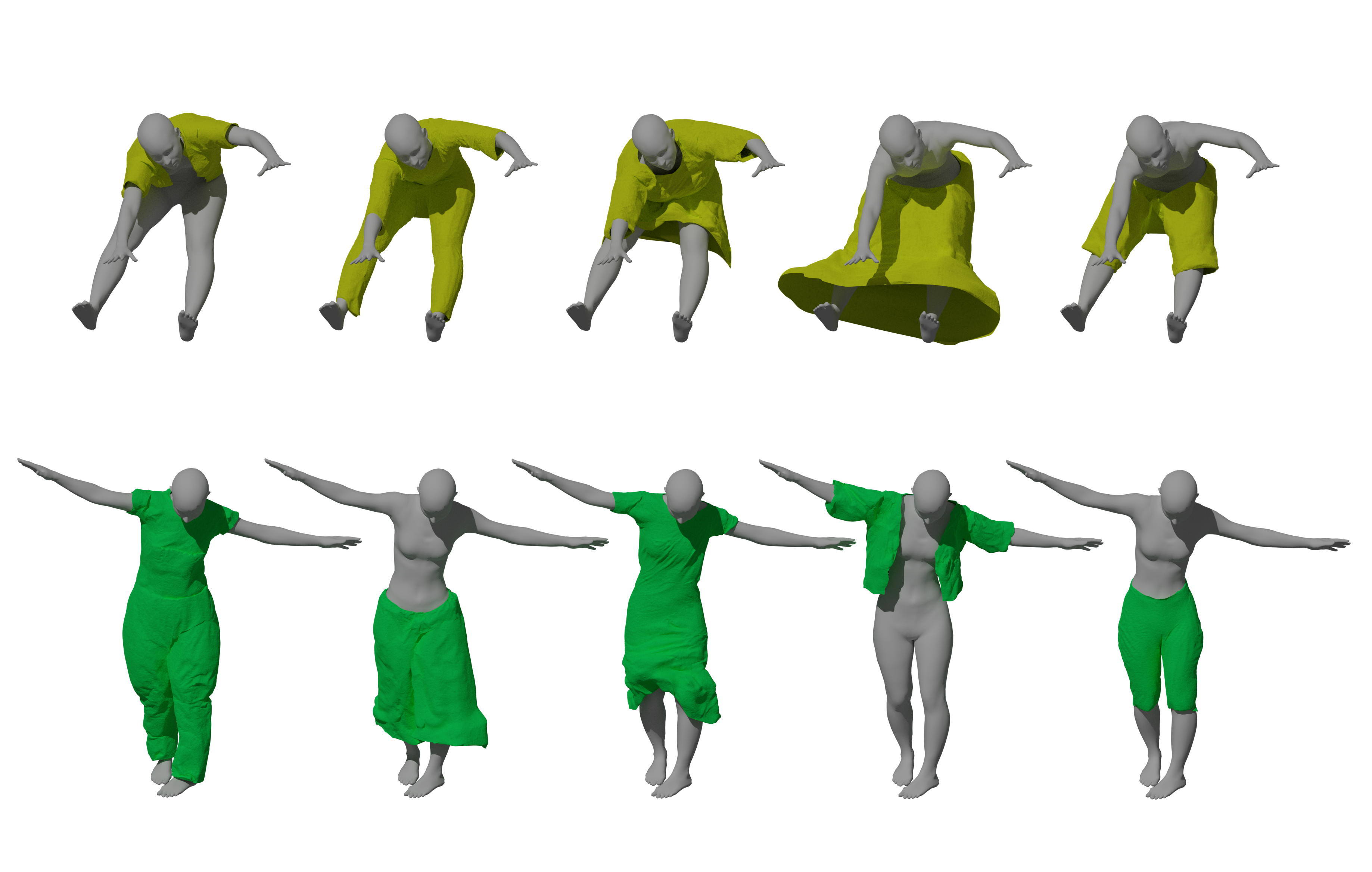}
        \caption{Garment variation.}
        \label{fig:trans_static}
    \end{subfigure}
    \begin{subfigure}{0.48\textwidth}
        \includegraphics[width=\textwidth]{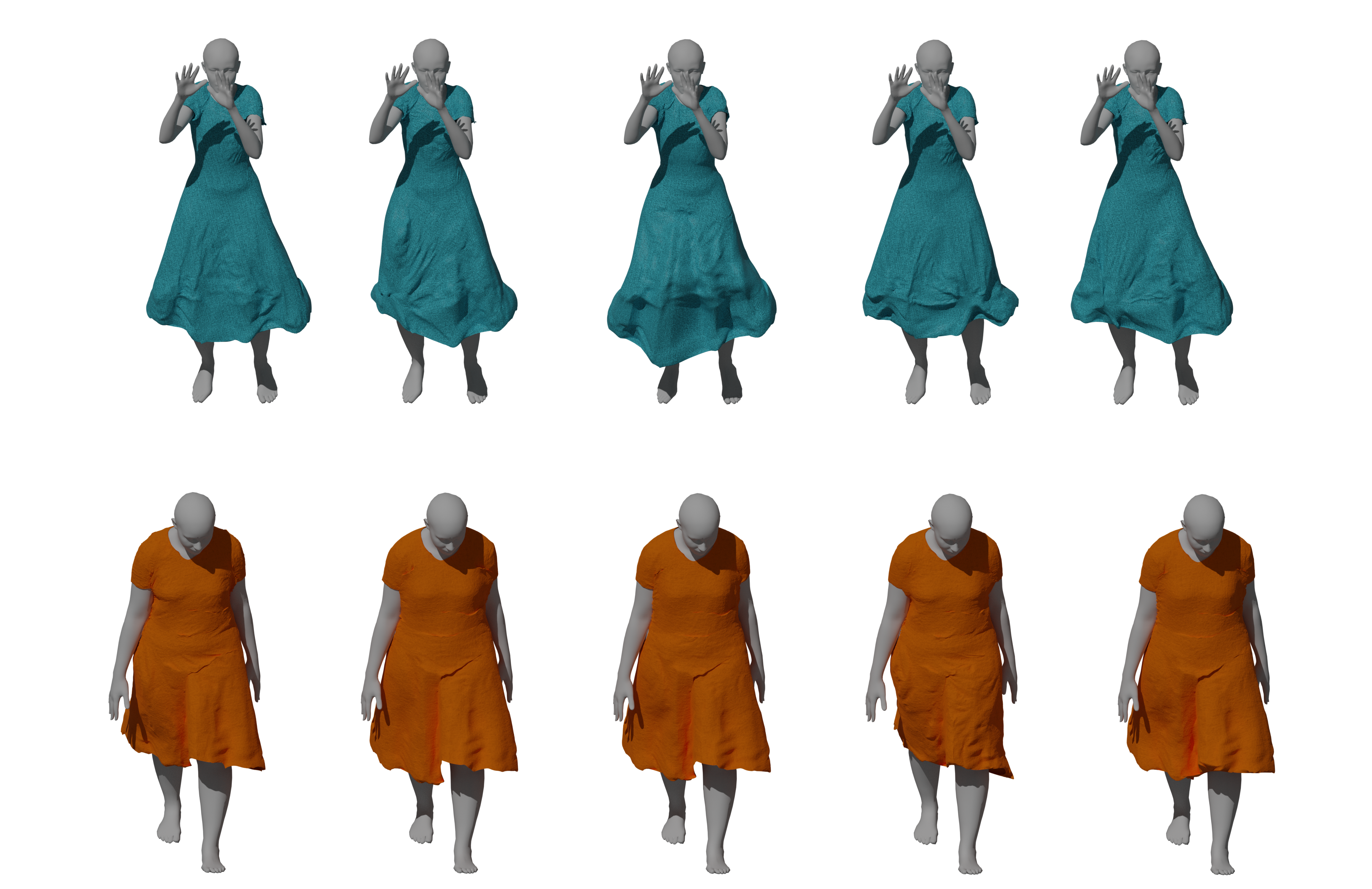}
        \caption{Wrinkles variation.}
        \label{fig:trans_dynamic}
    \end{subfigure}
    \begin{subfigure}{0.45\textwidth}
        \includegraphics[width=\textwidth]{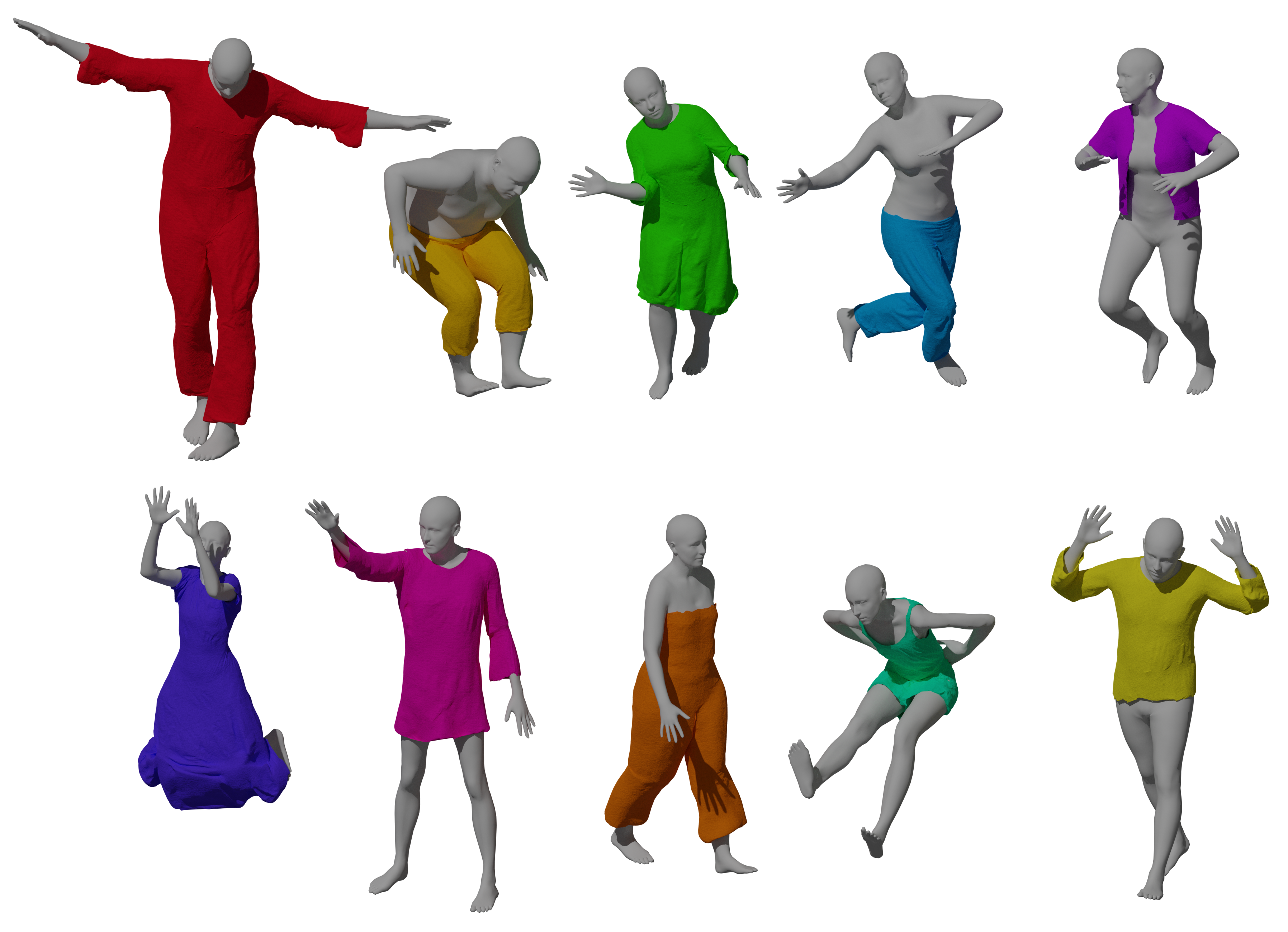}
        \caption{All variations.}
        \label{fig:trans_all}
    \end{subfigure}
    \caption{Generated samples from learnt latent codes, conditioning on different variables for dynamic garments (DVAE network).}\label{fig:trans}
\end{figure*}

\subsection{Wrinkle factorization vs. the rest conditional variables}
\label{sec: factorization}
We introduce two versions of DVAE. In one version, we computed offsets $X$ as $C-T$ and concatenated with $T$ as input to the network (as explained in the main paper). In this version, offsets are not invariant to pose. To better factorize $z_d$ to code garment wrinkles, we also implemented another version of DVAE where offsets are partially invariant to pose and garment type. Let $\mathcal{K}(A, \theta, J)$ be forward kinematic function that receives articulated object $A\in\mathbb{R}^{V_A\times3}$ with joints $J\in\mathbb{R}^{V_J\times3}$ and transforms it w.r.t. the rotations (or pose) $\theta$. Note that $A$ and $J$ are in rest pose in this definition. Here, $A$ can be replaced by $C$ or $T$. Since we register garments on top of body, garments can be transformed by body joints. Given this definition, $\mathcal{K}^{-1}$ can be defined as the inverse of the kinematic function which transforms back the object to its rest pose.

We define new offsets as $X^{new}=\mathcal{K}^{-1}(C_d, \theta, J) - C_s$ where $C_d$ is the dynamic dressed garment in pose $\theta$ and $C_s$ is the static garment in rest pose. We then concatenate $X^{new}$ with $C_s$ and $T$, normalize it and feed the network. Finally, at the output of the network, garments are reconstructed by the reverse process. Since we train SVAE to reconstruct $C_s$, we can rely on it at test time to generate dressed garments. We train this network with the same loss functions as before. By doing this factorization on the offsets, we are able to feed the network with relevant information of the wrinkles and factorize $z_d$ in a more meaningful way.

\subsection{Wrinkles factorization results}
\label{sec: results}
In Fig. \ref{fig:trans} we show some qualitative results of the learnt latent space and conditioning on different variables, specifically pose ($\theta$), garment code ($z_s$) and wrinkle code ($z_d$). One can see the network can learn a meaningful and consistent space. Regarding the learnt wrinkle code (Fig. \ref{fig:trans_dynamic}), a rest pose (upper row) shows less wrinkle variability than a complex action category (lower row). This is while by conditioning on pose (Fig. \ref{fig:trans_pose}) or garment code (Fig. \ref{fig:trans_static}), we can accurately retarget fixed wrinkle codes to new scenarios.

\subsection{Dynamic Garment Generation error.}
\label{sec:dyn}
\begin{figure}
    \centering
    \includegraphics[width=.65\textwidth]{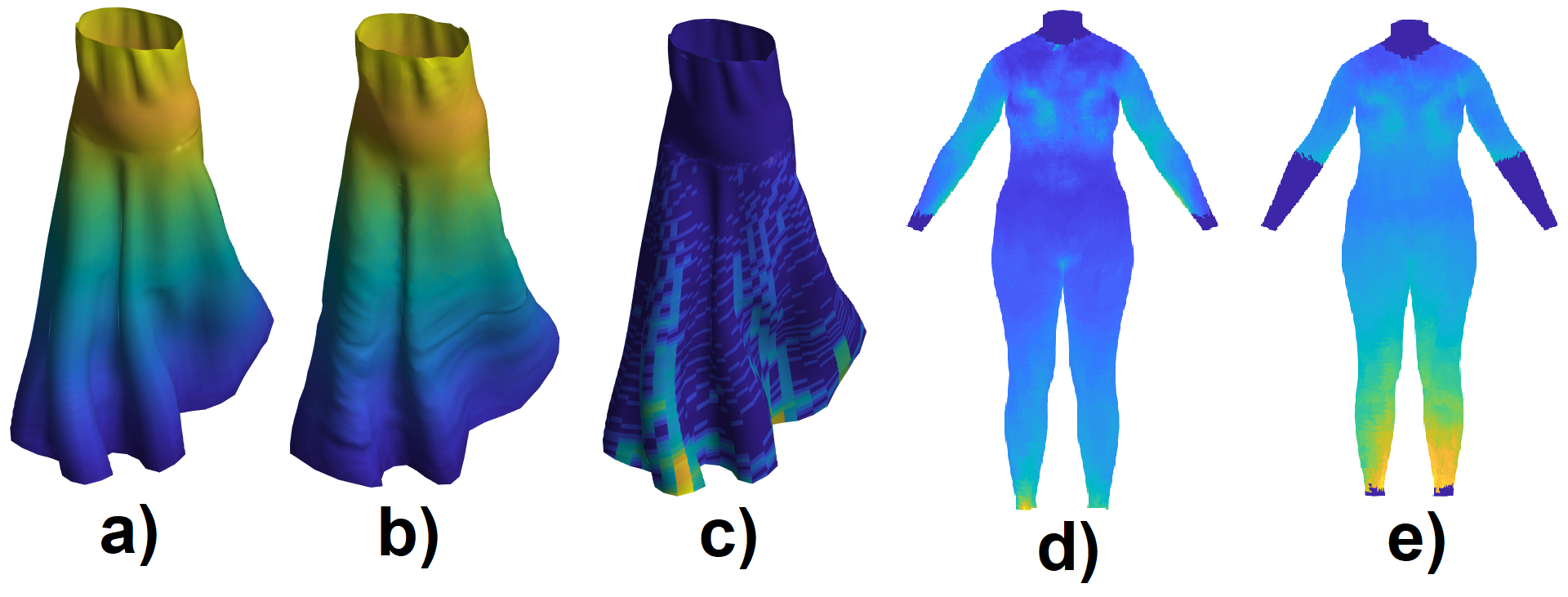}
    \caption{Implicit registration and S/DVAE per vertex error. a) original garment. b) reconstruction from offsets and body vertices. c) original vs. reconstruction error heatmap (scale in cm.). d-e) SVAE vs. DVAE garment generation error heatmaps. Dark blue=0, yellow$>$10cm. }
    \label{fig:reconstruction_error}
\end{figure}

As explained in the main paper, this work proposes encoding garments into SMPL body (subdivided for higher resolution) as offsets. This creates and implicit registration error, we represent its effects on Fig.\ref{fig:reconstruction_error}a-c. Additionally, we analyze the error distribution through the body. Per vertex reconstruction error is shown in Fig. \ref{fig:reconstruction_error} for static and dynamic samples. It can be seen the error is higher near the feet. This is because of the dynamics of skirts and dresses, which have the highest error (see Tab.4 on the main paper). Interestingly, trousers have the lowest error among others.

\subsection{Post-processing}
\label{sec: post}
\begin{figure}
    \centering
    \includegraphics[width=0.95\textwidth]{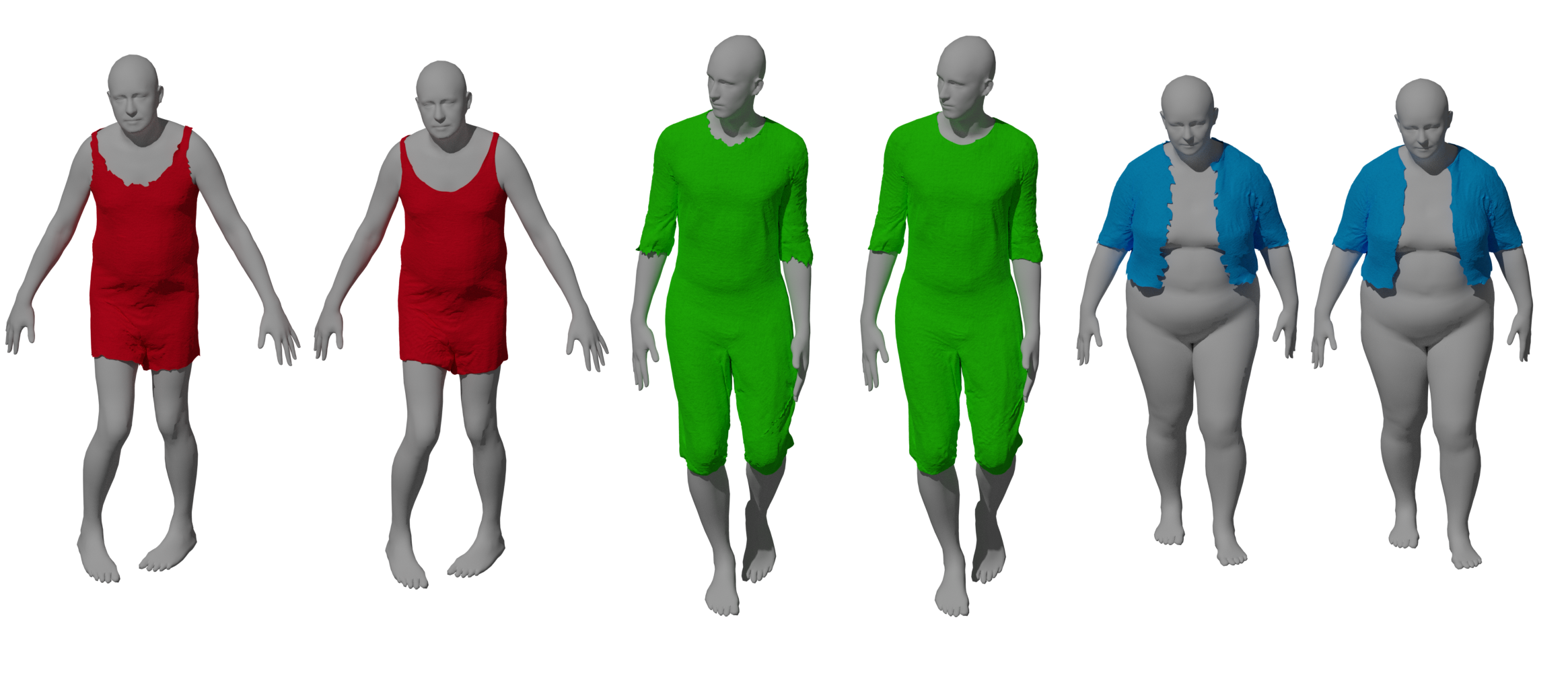}
    \caption{Boundaries are noisy as mask prediction is noisy at boundaries as well. With the proposed post-processing we can mitigate most of this noise, yielding visually appealing garments. Left: raw prediction. Right: post-processing results.}
    \label{fig:post}
\end{figure}

Since garments are encoded as SMPL offsets, their meshes follow the same connectivity of vertices as the human body model, except for dresses and skirts which follow the topology proposed at Sec. 4.1 on the main paper. Then, garment vertices are selected according to the predicted mask. In practice, this yields noisy boundaries. Fortunately, this is easily solvable through a simple post-processing. Garment's boundary vertices are identified by checking their neighbour's mask prediction. A normalized Laplacian matrix $\mathbf{L}\in \mathbb{R}^{14475\times14475}$ is constructed such that it represents a new connectivity where only boundary vertices are linked by edges. Then, smoothing boundaries can be achieved by iteratively multiplying the garment vertices $\mathbf{X}\in\mathbb{R}^{14475\times3}$ as: 
\begin{equation}
    \mathbf{X}_{i+1} = \mathbf{L} \mathbf{X}_{i},
\end{equation}
where $i$ is the iteration count. Empirically, $10$ iterations are more than enough to produce visually appealing results, while increasing iterations might destroy high-frequency details. This smoothing can be efficiently computed as sparse matrix multiplication. In Fig.\ref{fig:post} are depicted a few samples before and after applying the described post-processing. We observe how the proposed method can improve noisy boundaries and generate more natural-looking garments.

\section{Applications of the Dataset}
\label{sec:applications}
\begin{figure}
    \centering
    \includegraphics[width=.75\linewidth]{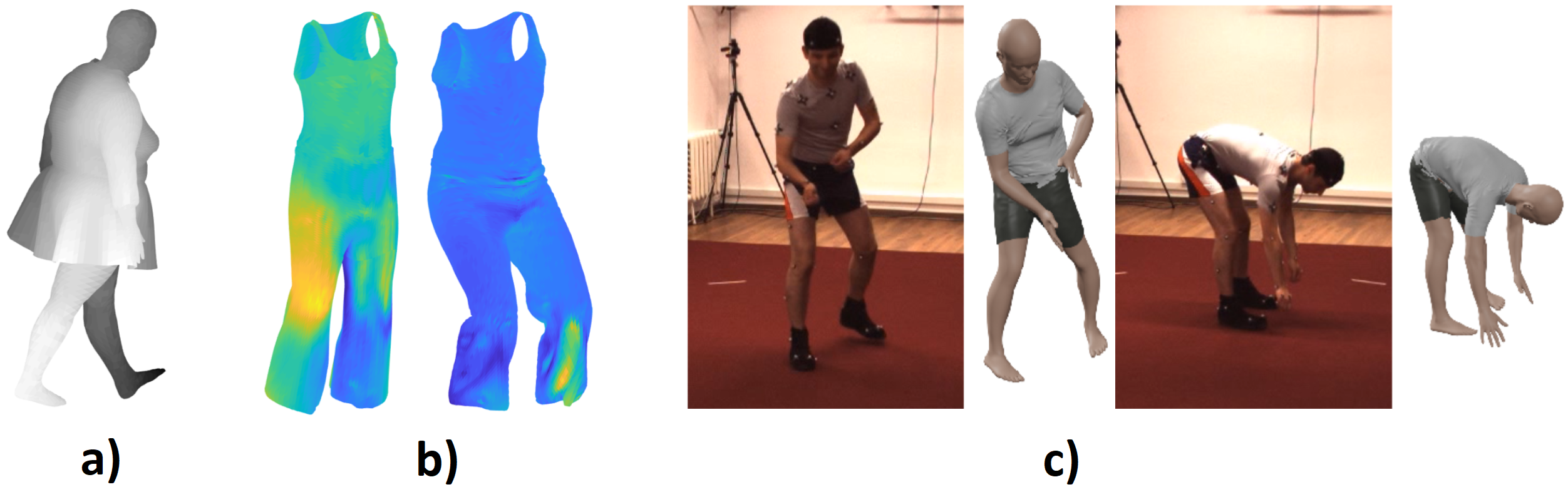}
    \caption{Applications of CLOTH3D. a): A rendered depth image, b): vertex velocities (Dark blue=0, yellow$>$0.8.), c): Automatic RGB to 3D cloth estimation.}
    \label{fig:app}
\end{figure}

CLOTH3D could be used not just for 3D garment generation but in other application scenarios, such as human pose and action recognition in depth images, garment motion analysis, filling missing vertices of scanned bodies with additional meta data (e.g. garment segments), support  designers and animators tasks, or estimating 3D garment from RGB images, just to mention a few. We ran some proof-of-concept applications using our CLOTH3D data, shown in Fig. \ref{fig:app}: a rendered depth image, garment motion velocities, and RGB-to-3D cloth estimation. For the later, given our layered garment structure and SMPL segmentation, we rendered 10K samples of t-shirts and trousers with different poses of Human3.6M~\cite{h36m_2014} dataset. These data contains images of body segments and garment silhouette. We then trained ResNet50 to regress available static garment codes. In test time, we assume body shape, pose and garment silhouette are available. This information can be extracted by state-of-the-art SMPL based pose estimation and cloth parsing methods. In our case, we manually segmented garments of two frames of this dataset (shown in Fig. \ref{fig:app}(c)) and used them to estimate garment code. We then copied the wrinkles from the nearest sample in CLOTH3D. Finally 3D garments were reconstructed and rendered as shown.

\end{document}